\pdfoutput=1

\documentclass[11pt]{article}

\usepackage[final]{acl}

\usepackage{times}
\usepackage{latexsym}

\usepackage[T1]{fontenc}

\usepackage[utf8]{inputenc}

\usepackage{microtype}

\usepackage{inconsolata}

\usepackage{graphicx}

\usepackage{amsmath}
\usepackage{amsfonts} 
\usepackage{amssymb}
\usepackage{amsthm}
\usepackage{mathrsfs}
\usepackage{mathtools} 
\usepackage{tabularx}
\usepackage{booktabs}
\usepackage{multicol}
\usepackage{multirow}
\usepackage{caption}
\usepackage{graphicx}
\usepackage{float} 
\usepackage{subfigure}
\usepackage{subcaption}
\usepackage{soul} 
\usepackage{enumitem}
\usepackage{hyperref}

\newcommand{\tabincell}[2]{\begin{tabular}{@{}#1@{}}#2\end{tabular}}

\definecolor{light_blue}{HTML}{68bed9}
\definecolor{orange}{HTML}{ed8d5a}

\title{Optimal Transport-Based Token Weighting scheme \\ for Enhanced Preference Optimization}

\author{
 Meng Li\textsuperscript{1}\footnotemark[1],
 Guangda Huzhang\textsuperscript{2},
 Haibo Zhang\textsuperscript{2},
 Xiting Wang\textsuperscript{1}\footnotemark[2]\footnotemark[3],
 Anxiang Zeng\textsuperscript{2}\footnotemark[2]
\\
 \textsuperscript{1}Gaoling School of Artificial Intelligence, Renmin University of China, \\
 \textsuperscript{2}LLM Team, Shopee Pte. Ltd.
\\
 \texttt{mengli.24@ruc.edu.cn, guangda.huzhang@shopee.com, peter.wu@shopee.com}\\
 \texttt{xitingwang@ruc.edu.cn, zeng0118@e.ntu.edu.sg} 
}

\begin{document}
\maketitle

\renewcommand{\thefootnote}{\fnsymbol{footnote}} 
\footnotetext[1]{Work done during internship at Shopee LLM Team.} 
\footnotetext[2]{Corresponding Authors.} 
\footnotetext[3]{Work partly done at Beijing Key Laboratory of Research on Large Models and Intelligent Governance and Engineering Research Center of Next-Generation Intelligent Search and Recommendation, MOE.} 
\renewcommand{\thefootnote}{\arabic{footnote}}

\begin{abstract}

Direct Preference Optimization (DPO) has emerged as a promising framework for aligning Large Language Models (LLMs) with human preferences by directly optimizing the log-likelihood difference between chosen and rejected responses. However, existing methods assign equal importance to all tokens in the response, while humans focus on more meaningful parts. This leads to suboptimal preference optimization, as irrelevant or noisy tokens disproportionately influence DPO loss. To address this limitation, we propose \textbf{O}ptimal \textbf{T}ransport-based token weighting scheme for enhancing direct \textbf{P}reference \textbf{O}ptimization (OTPO). By emphasizing semantically meaningful token pairs and de-emphasizing less relevant ones, our method introduces a context-aware token weighting scheme that yields a more contrastive reward difference estimate. This adaptive weighting enhances reward stability, improves interpretability, and ensures that preference optimization focuses on meaningful differences between responses. Extensive experiments have validated OTPO's effectiveness in improving instruction-following ability across various settings.\footnote{Code is available at \href{https://github.com/Mimasss2/OTPO}{https://github.com/Mimasss2/OTPO}.}

\end{abstract}
\section{Introduction}
\label{sec:intro}

Aligning large language models with human preferences~\cite{ouyang2022training} and values~\cite{yao2023instructions,yi2023unpacking} guides LLMs to be helpful, honest, and harmless, preventing misuse of their powerful abilities~\cite{bai2022training}. Reinforcement Learning from Human Feedback (RLHF) achieves this objective via fine-tuning the LLM to optimize the learned reward model~\cite{ouyang2022training}. 
Offline direct preference optimization algorithms, e.g., DPO~\cite{rafailov2024direct}, simplify this process by applying reparameterization to implicitly model the reward as the log ratio likelihood of the response, which is equivalent to the sum of log ratio likelihoods of all tokens. This transformation results in a simple binary cross-entropy objective of reward difference, and has been widely adopted due to its training stability and efficiency~\cite{xiao2024comprehensive}.\looseness=-1

\begin{figure}[t]
    \centering
    \includegraphics[width=0.98\linewidth]{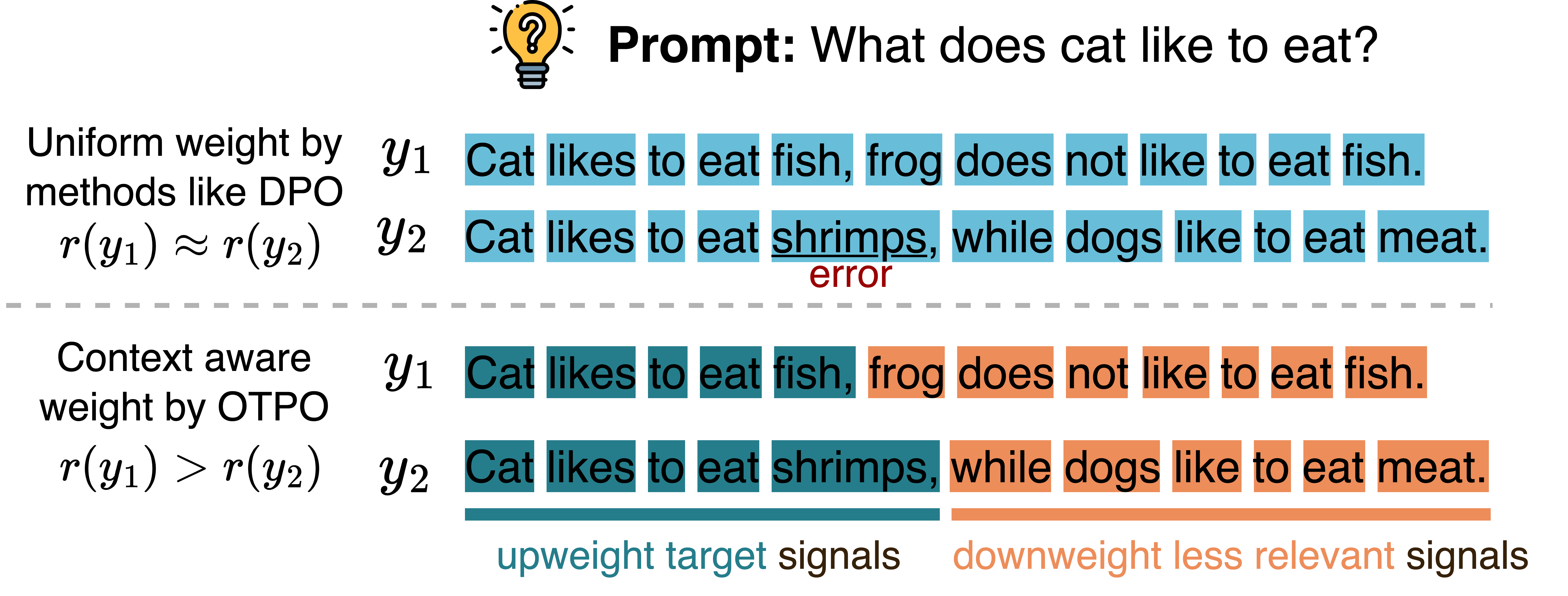}
    \caption{The uniform weighting in DPO leads to suboptimal alignment results, allowing less relevant signals to dominate. OTPO identifies the contextually similar parts in pairwise responses as targets and upweights target signals. $r(y_*)$ denotes the estimated reward under each method.}
    \label{fig:comparison_figure}
    \vspace{-15pt}
\end{figure}

The DPO loss treats each token equally, which can bias the model to overlook less important factors and learn by shortcuts, leading to suboptimal results~\cite{park2024disentangling}. In Fig.~\ref{fig:comparison_figure}, tokens less relevant to the question dominate the reward, while important parts like ``\textit{Cat likes to eat fish}'' should be paid more attention.
Current methods, including SimPO~\cite{meng2024simpo}, SamPO~\cite{lu2024eliminating}, and LDDPO~\cite{liu2024length}, primarily focus on the length bias caused by imbalanced total token weight between a chosen response and a rejected response. They apply a heuristic weighting scheme to reduce the difference in total token weight. Moreover, they can not distinguish the important tokens relevant to instruction-following due to a lack of supervision signal. Recent work like APO~\cite{dao2024flashattention} has attempted to address this issue by rewriting the irrelevant parts to maintain minimum difference with the other response, yet it relies heavily on the external reviser.

In this paper, we propose an \textbf{O}ptimal \textbf{T}ransport-based weighting scheme for direct \textbf{P}reference \textbf{O}ptimization (OTPO), a novel unsupervised framework for calculating token weights in direct preference optimization.
Our key innovation lies in emphasizing tokens where the responses agree (similar tokens) as indicators of higher quality or shared information, and de-emphasizing tokens where they disagree.  
We observe that these shared parts of responses are more likely to be relevant to the question, as there are multiple ways to represent the same answer. 
Specifically, we utilize an unbalanced optimal transport approach to dynamically assign a fixed total weight budget to token-level weights based on the similarity between tokens in chosen and rejected responses, allocating higher weights to more semantically relevant tokens. This allows for estimating the minimum effort required to transform one response to the other. Compared to previous methods, our method improves reward stability, enhances the transparency of preference optimization, and ensures the optimization focuses on more meaningful differences between responses.

To sum up, our contributions are threefold:
\begin{itemize}
    \item We identify the issue of treating tokens equally in DPO and propose a general weighting scheme that incorporates previous methods.
    \item We design an optimal transport-based token weighting scheme to identify important tokens without extra supervision signal.
    \item Extensive experiments have validated OTPO's effectiveness across various settings, achieving up to 10.9\% length-controlled win rate increase on AlpacaEval2 compared to DPO. 
\end{itemize}

\section{Methods}

\begin{figure*}[t]
    \centering
    \includegraphics[width=0.99\linewidth]{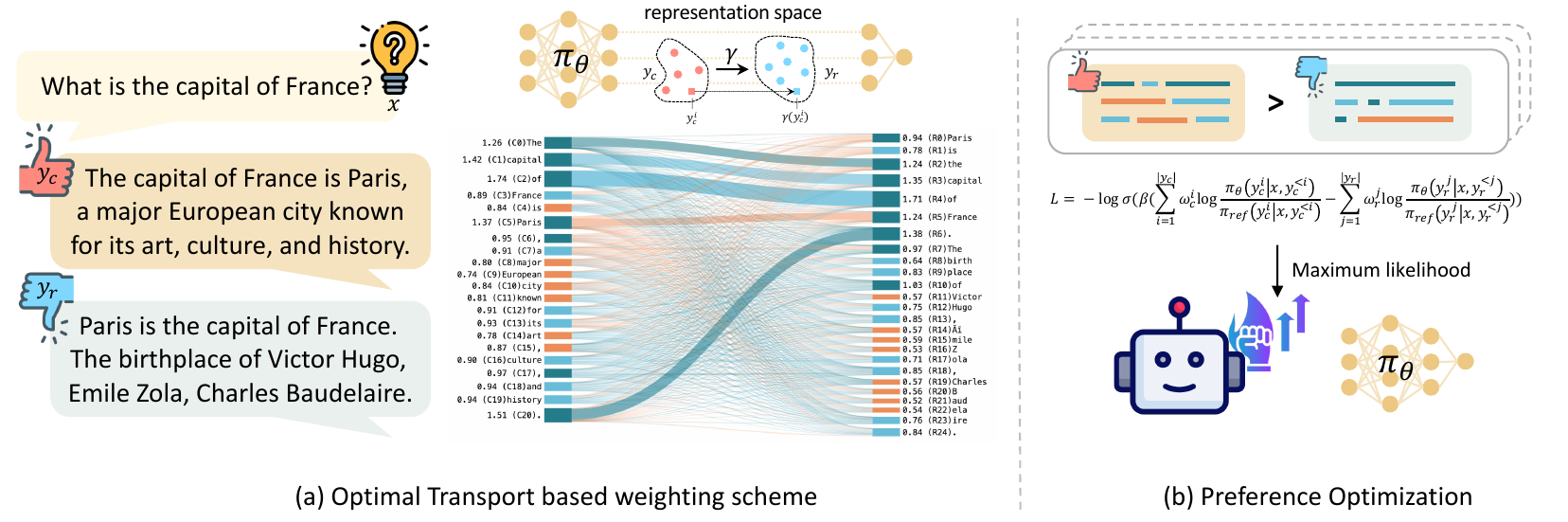}
    \caption{Overall framework. (a) We compute the token-level weighting scheme using optimal transport. Each response's distribution is made up of its tokens, represented as vectors in the LLM's representation space. The optimized transport plan is visualized using a Sankey diagram. (b) We decompose the DPO loss at the token level and apply the weighting scheme obtained in (a).  }
    \label{fig:teasor}
    \vspace{-10pt}
\end{figure*}

We first provide a simple background of DPO (Sec.~\ref{sec:dpo_background}).
To refine the understanding of the reward difference term $\Delta_r$ in DPO, we decompose it from a more granular perspective and propose a general weighting scheme that incorporates previous methods (Sec.~\ref{sec:decompose_dpo}). By examining interactions at the level of chosen-rejected token pairs, we identify opportunities for improvement in how reward differences are computed and propose our OTWPO algorithm for more nuanced adjustments (Sec.~\ref{sec:ot_weight}). Fig.~\ref{fig:teasor} shows the overall framework.

\subsection{Background: Direct Preference Optimization}
\label{sec:dpo_background}

DPO eliminates the need for explicitly learning a reward model and reparameterizes the reward model as:
\begin{equation}
    r(x,y) = \beta \log \frac{\pi_{\theta}(y|x)}{\pi_{\text{ref}}(y|x)} + \beta \log Z(x)
    \label{eq:dpo_reward}
\end{equation}
Here, $\pi_{\theta}$ is the model to be optimized, $\pi_{\text{ref}}$ is a reference model, $\pi_{*}(y|x)$ denotes the probability of a response $y$ given input $x$ under policy $\pi_{*}$, and $Z(x)$ is an unknown partition function. 
Incorporating the above reward model into the Bradley-Terry model, the final DPO loss function is: 
\begin{align}
    \mathcal{L}(\pi_{\theta}; \pi_{\text{ref}}) = - \mathbb{E}_{(x, y_c, y_r) \sim D} 
    [ \log \sigma(\beta \Delta_r)]
    \label{eq:dpo_loss}
    \\
    \Delta_r = \log\frac{\pi_{\theta}(y_c|x)}{\pi_{\text{ref}}(y_c|x)} 
    -  \log\frac{\pi_{\theta}(y_r|x)}{\pi_{\text{ref}}(y_r|x)}
    \label{eq:reward_delta}
\end{align}
where $(x, y_c, y_r)$ is a preference pair consisting of a prompt $x$, a chosen response $y_c$, and a rejected response $y_r$ from the preference dataset $D$. And $\sigma$ denotes the sigmoid function.

Formally, given a response $y$ of length $|y|$, its probability under the policy is factorized as the multiplicative product of each token's probability $\pi_{\theta}(y|x) = \prod_{i=1}^{|y|}\pi_{\theta}(y^i|x, y^{<i})$. The reparameterized reward difference term $\Delta_{r}$ in DPO treats the entire response as a single action, in contrast to classical RLHF methods that model each token as an action and optimize token-level value functions with sparse rewards at the terminal state~\cite{rafailov2024r}.
This can mislead optimization by causing the policy to focus on less important tokens and learn through shortcuts, potentially undermining the intended reward signal.

\subsection{Decomposing DPO Loss}
\label{sec:decompose_dpo}

We first break down the reward difference $\Delta_r$ in DPO loss at the token level to explicitly reveal how each token contributes to the optimization process. Operating at the token level, we have (see proofs in Appx.~\ref{app:token_dpo_loss}):
\begin{equation}
    \begin{split}
    \Delta_r = \sum_{i=1}^{|y_c|}  q_c^i - \sum_{j=1}^{|y_r|} q_r^j, \\
    \text{where} \quad
    q_*^i = \log\frac{\pi_{\theta}(y_*^i|x, y_*^{<i})}{\pi_{\text{ref}}(y_*^i|x, y_*^{<i})}
    \end{split}
    \label{eq:reward_diff_token}
\end{equation}
Here, $q_*^i$ denotes the log-likelihood ratio of the $i$-th token in $y_*$ between the model and the reference distribution. We can see that each token contributes equally by its log-likelihood ratio. We further incorporate a token-level weighting scheme for $\Delta_r$ and express it as:
\begin{equation}
    \Delta_r = \sum_{i=1}^{|y_c|} \omega_c^i  q_c^i - \sum_{j=1}^{|y_r|} \omega_r^j q_r^j
    \label{eq:weighted_reward_diff}
\end{equation}
where $\omega_*^i$ represents the weight assigned to the $i$-th token in response $y_*$. This decomposition expresses $\Delta_r$ in terms of differences in weighted token log-probability ratio difference. DPO can be viewed as a special case of Eq.~\ref{eq:weighted_reward_diff}, assigning a uniform weight of 1 for all tokens.

\begin{table}[t!]
\centering
\begin{tabular}{cc}
    \toprule
    \textbf{Method} & \textbf{Weighting Scheme}  \\
    \midrule
    \textbf{DPO} & $\omega_i = 1, \, \forall i \in [1, |y|]$  \\
    \midrule
    \textbf{SimPO} & $\omega_i = \frac{1}{|y|}, \, \forall i \in [1, |y|]$  \\
    \midrule
    \textbf{SamPO} & 
    \begin{tabular}[c]{@{}c@{}}
    $\omega_i = 1, \, \forall i \in S$ , \; 
    $\omega_i = 0, \, \forall i \notin S$ \\ 
    where $S \sim \text{Uniform}(m, [1, |y|])$
    \end{tabular}  \\
    \midrule
    \textbf{LDDPO} & 
    \begin{tabular}[c]{@{}c@{}}
    $\omega_i = 1, \, \forall i \in [1, m]$ \\ 
    $\omega_i = \alpha, \, \forall i \in [m+1, |y|]$
    \end{tabular} \\
    \midrule
    \textbf{OTPO} & 
    \begin{tabular}[c]{@{}c@{}}
    $\omega_c^i = \sum_{j=1}^{|y_r|} \Gamma_{i,j}$ , \;
    $\omega_r^i = \sum_{i=1}^{|y_c|} \Gamma_{i,j}$
    \end{tabular} \\
    
    \bottomrule
\end{tabular}
\caption{Weighting schemes for different methods. Here, $|y|$ is the current response length, $|y_c|$ and $|y_r|$ denote the lengths of the chosen and rejected responses, respectively. $m = \min(|y_c|, |y_r|)$, and $\alpha \in [0,1]$ is a hyperparameter in LDDPO. $\Gamma$ is the optimized transport plan in Eq.~\ref{eq:po_transport_plan}.}
\label{tab:weight_baselines}
\vspace{-15pt}
\end{table}

\begin{figure*}
    \centering
    \includegraphics[width=0.99\linewidth]{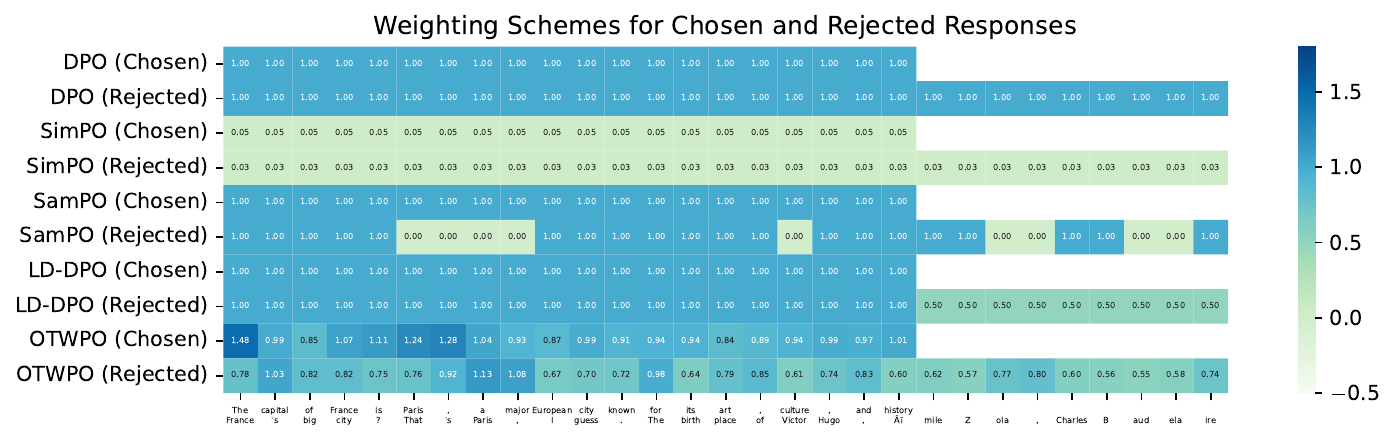}
    \caption{Weights assigned to the responses given different methods. Here, given the prompt \textit{``What is the capital of Paris?''}, the chosen response is \textit{``The capital of France is Paris, a major European city known for its art, culture, and history.''}, and the rejected response is \textit{``France’s big city? That’s Paris, I guess. The birthplace of Victor Hugo, Émile Zola, Charles Baudelaire.''}}
    \label{fig:weight_baselines}
    \vspace{-15pt}
\end{figure*}

This token weighting scheme incorporates previous methods for mitigating length bias, as shown in Tab.~\ref{tab:weight_baselines}. Fig.~\ref{fig:weight_baselines} provides a more intuitive illustration. Length bias refers to the phenomenon where the model learns to only improve length instead of quality to increase reward difference compared to the base model. 
DPO causes length bias as the difference in the two responses' total token weight $\delta = \sum_{i=1}^{|y_c|} \omega_c^i - \sum_{j=1}^{|y_r|} \omega_r^j = |y_c| - |y_r|$ being positive most of the time. Previous methods essentially apply a weighting scheme to reduce the total token weight difference $\delta$. SimPO down-weight all tokens' weight to $1/|y|$, ensuring each response's total weight sums to 1. SamPO employs a more subtle downsampling on the longer response to only consider the same amount of random tokens as the other response. LDDPO only down-weights the over-lengthy parts to reduce $\delta$. These methods apply some heuristic kind of weighting scheme to mitigate length bias caused by total token weight bias. Yet a more principled weighting scheme delving into each token's importance is needed to solve the problem fundamentally.

\subsection{Optimal Transport based Weighting Scheme}
\label{sec:ot_weight}

While prior methods primarily adjust total token weights, our approach takes a deeper look into the geometric structure of token pair relationships. We further break down the reward difference term as the weighted token log-likelihood difference across all chosen-rejected token pairs:
\begin{equation}
    \begin{split}
        \Delta_r = \sum_{i}\sum_{j} \Gamma_{i,j} (q_c^i - q_r^j)
    \end{split}
    \label{eq:token_pair_delta_r}
\end{equation}
Here, $\Gamma_{i,j}$ represents the weight assigned to each token pair $\{y_c^i, y_r^j\}$. Then, the previous token level weights $\omega_c^i, \omega_r^j$ corresponds to:
\begin{equation}
    \omega_c^i = \sum_{j} \Gamma_{i,j}, \quad \omega_r^j = \sum_{i} \Gamma_{i,j}.
        \label{eq:token_weights}
\end{equation}

Building on the token-pair level transformation of the DPO loss, we now focus on exploiting this finer granular $\Delta_r$ to differentiate token pairs based on their semantic relevance.
We aim to emphasize semantically meaningful token pairs, with similar meanings and structural roles, while de-emphasizing less relevant ones.
This coincides with the principle of majority voting, where the most frequently occurring element is chosen as the final answer~\cite{wang2023selfconsistency}. In our context, the parts of the responses most relevant to the question—such as the direct answer ``The capital of France is xxx''—are more likely to appear consistently across both the chosen and rejected responses. This is particularly evident in the on-policy setting, where all responses are generated by the same policy. These shared parts, representing the ``majority'', carry critical information and thus are prioritized.

A key challenge arises from the unequal total weight sums of chosen and rejected responses in naive DPO loss, while the above formulation inherently requires the same total weight sum for each response pair, as:
\begin{equation}
    \sum_{i} \omega_c^i  = \sum_{j} \omega_r^j  = 
    \sum_{i} \sum_{j} \Gamma_{i,j}
\end{equation}
The optimal $\Gamma$ is computed by solving the optimization problem later defined in Equation 9. We ideally want a weighting scheme that accounts for their structural differences while still ensuring fairness in total weight.
A natural way to achieve this is through optimal transport, which provides a principled method for aligning distributions while minimizing their discrepancy. However, standard optimal transport assumes equal total mass on both sides, making it incompatible with our setting. To address this, we adopt unbalanced optimal transport, which allows for flexible mass redistribution between the two responses while preserving meaningful semantic differences.

The cornerstone of OTPO lies in constructing a cost matrix $M \in \mathbb{R}^{|y_c| \times |y_r|}$, where each entry $M_{ij}$ quantifies the distance between the $i$-th token of the chosen response $y_c$ and the $j$-th token of the rejected response $y_r$.  
Since the optimal transport framework requires the transported distribution to reside in a proper metric space, directly using the log-likelihood ratio difference in Eq.~\ref{eq:token_pair_delta_r} is not feasible, as it does not naturally form a metric space. We take a step back and leverage the last-layer token representations, which better preserve the underlying semantic structure.
We specifically use euclidean distance, i.e., $M_{ij} = \| h_c^i - h_r^j \|_2$,  as it is commonly used in metric space~\cite{arjovsky2017wasserstein}.
Here, $h_*^t$ is the hidden state of the $t$-th token in response $y_*$, extracted from the model's hidden representation space.

Building on this cost matrix, we define the optimization problem as:  
\begin{equation}
    \begin{split}
        \Gamma^* = \arg \min_{\Gamma } \sum_{i,j} \Gamma_{i,j} M_{i,j} + \epsilon_1 \Omega(\Gamma) \\
    + \epsilon_2 (\mathbb{KL}(\Gamma \mathbf{1}, \mathbf{1}_{|y_c|}) + \mathbb{KL}(\Gamma ^T\mathbf{1}, \mathbf{1}_{|y_r|}))
    \end{split}
    \label{eq:po_transport_plan}
\end{equation}
where $\Gamma \in \mathbb{R}^{|y_c| \times |y_r|}$ represents the transport plan that aligns tokens between the chosen and rejected responses.  
$\Omega(\Gamma) = \sum_i \sum_j \Gamma_{i,j}\log(\Gamma_{i,j})$ is an entropy regularizer, controlling the sparsity of the transport plan. 
Meanwhile, the $\mathbb{KL}(\cdot)$ terms ensure that the marginal distributions of $\Gamma$ are close to the naive DPO uniform weights, allowing for controlled deviations. This formulation unifies semantic alignment and token weight control under an optimal transport framework, whereas the first term corresponds to the Wasserstein distance between the two responses' distribution.

After solving for the optimal transport plan $\Gamma^*$, the token-level weights $\omega_c^*$ and $\omega_r^*$ are obtained by summing along the respective dimensions as in Eq.~\ref{eq:token_weights} and normalizing to a predefined scale $\tau$ to ensure optimization stability:
\begin{equation}
    \omega_c^* = \frac{\Gamma \mathbf{1}}{|\Gamma|} \tau, \quad \omega_r^* = \frac{\Gamma^\top \mathbf{1}}{|\Gamma|} \tau
    \label{eq:rescale}
\end{equation}
Here, $|\Gamma|$ represents the total weight sum of the transport plan. The normalized transport plan is equal to the one in Eq.~\ref{eq:po_transport_plan} when down-weighting the distance term by $\tau/|\Gamma|$. This allows for automatic total weight adjustment based on response length, therefore leading to a more stabilized total weight budget and preference optimization. We specifically set $\tau = \min(|y_c|, |y_r|)$ to ensure the total weight corresponds to the public length, 
which is enough to contain a more concise representation of relevant information most of the time. 
This choice helps balance the contributions of the two responses and reduces the disproportionate influence of less relevant tokens.

Then we incorporate the optimal token-level weights $\omega_c^*$ and $\omega_r^*$ in Eq.~\ref{eq:rescale} into the reward estimation, replacing the uniform weights used in standard DPO. This allows the model to focus on semantically significant tokens, yielding a reward difference estimate:
\begin{equation}
    \begin{split}
        \Delta_{\hat{r}} = \sum_{i=1}^{|y_c|} \omega_c^{*i} \log \frac{\pi_\theta(y_c^i|x, y_c^{<i})}{\pi_{\text{ref}}(y_c^i|x, y_c^{<i})} \\
        - \sum_{j=1}^{|y_r|} \omega_r^{*j} \log \frac{\pi_\theta(y_r^j|x, y_r^{<j})}{\pi_{\text{ref}}(y_r^j|x, y_r^{<j})}
    \end{split}
\end{equation}
This weighted reward difference captures the fine-grained contributions of individual tokens to the overall preference.
The final OTPO loss is formulated as:
\begin{equation}
    \mathcal{L}(\pi_\theta; \pi_{\text{ref}}) = - \mathbb{E}_{(x, y_c, y_r) \sim D} \left[\log \sigma(\beta \Delta_{\hat{r}})\right]
\end{equation}

In summary, OTPO leverages Optimal Transport to dynamically assign a fixed weight budget to token pairs based on their semantic relevance, enabling a fine-grained inspection of the reward difference term. This approach emphasizes meaningful token interactions while reducing the impact of less relevant or extraneous tokens, providing a more robust and interpretable optimization framework of LLM alignment.

\section{Experimental Setup}

We conduct preference optimization experiments to compare different optimization methods under various settings, including task, optimization strategy, and model. 
The tasks include general instruction-following and summarization. For general instruction-following, we conduct on-policy optimization on Llama-3-8B and Llama-3.2-3B, with UltraFeedback~\cite{cui2024ultrafeedback} and HelpSteer2~\cite{wang2024helpsteer2}'s preference version as the preference training dataset. For the summarization task, we perform offline optimization on Qwen-2.5-3B~\cite{yang2024qwen} using off-the-shelf TL;DR~\cite{stiennon2020learning} dataset.

\textbf{Model Training}.
In the general instruction-following task, we mainly consider off-the-shelf instruction-tuned models with more powerful instruction-following abilities. We first sample 10 responses per prompt for HelpSteer2 and 5 responses per prompt for UltraFeedback following~\cite{meng2024simpo,tunstall2023zephyr,wang2024helpsteer2}. Then, we annotate the sampled responses with ArmoRM~\cite{wang2024interpretable} and select the response with the highest and lowest score as $y_c, y_r$ respectively. 
In the summarization task, we first fine-tune \texttt{Qwen-2.5-3B} with the chosen responses, to obtain basic summarization capability, and then train the model directly with the TL;DR dataset.
We tune general hyperparameters using DPO and apply the set of hyperparameters to most of all preference optimization methods. Please refer to Appx.~\ref{app:implementation_details} for more detailed descriptions.

\textbf{Baselines.}
We primarily compare OTPO with DPO~\cite{rafailov2024direct} and other direct preference optimization methods, excluding RLHF approaches that require training an additional reward model, following prior works such as SimPO~\cite{meng2024simpo} and SamPO~\cite{lu2024eliminating}. Our focus is on methods that incorporate token-level weighting schemes. This includes SimPO, SamPO, and LDDPO~\cite{liu2024length}, which adjust the token weights to reduce total token weight differences and elevate length fairness. Please refer to Tab.~\ref{tab:weight_baselines} for more details and Fig.~\ref{fig:weight_baselines} for a more intuitive visualization of explicit token-level weighting schemes. We also include TDPO~\cite{zeng2024token}, which implicitly applies token weighting via token-level KL divergence. Additionally, we also include other variants of DPO, including length regularized DPO (LR-DPO), AOT~\cite{melnyk2024distributional}, Robust DPO~\cite{chowdhury2024provably}, RSO~\cite{liu2023statistical}, SPPO~\cite{wu2024self} in Appx.~\ref{app:evaluation_details}.

\textbf{Evaluation.}
We assess the alignment performance using GPT to perform pairwise comparisons.
We adopt AlpacaEval2~\cite{alpaca_eval,dubois2024length}, to assess the models' instruction-following ability. AlpacaEval2 contains 805 questions, and uses GPT-4-Turbo to perform side-by-side comparisons of the model response with a reference model (GPT-4 Preview). We report the win rate, length-controlled win rate, and response average length. The length-controlled win rate is computed by first estimating the impact of length differences on each test result and then adjusting for a length difference of zero to obtain a debiased estimation.
We specifically choose this benchmark, as it controls the effect of response length, while other LLM-as-a-judge benchmarks may exploit spurious correlations, including output length, presence of lists, position biases~\cite{zheng2023judging,koo2023benchmarking,wang2023large,wu2023style}. 
For the summarization task, we follow~\cite{rafailov2024direct} and evaluate the win rate against the base model, using GPT-4o as the judge model on 256 randomly sampled test cases from the TL;DR test set.
We also adopt MMLU, GSM8K, ARC Challenge, HellaSwag, and PiQA to examine the models' general ability on multiple domains in Appx.~\ref{app:evaluation_details}.

\section{Experimental Results}

\begin{table*}[!ht]
    \centering
    \begin{tabular}{cccccccc}
    \toprule
        ~ & \multirow{2}{*}{\textbf{Method}} & \multicolumn{3}{c}{\textbf{UltraFeedback}} & \multicolumn{3}{c}{\textbf{Helpsteer2}}  \\ 
        \cmidrule(lr){3-5} \cmidrule(lr){6-8}
        ~ & ~ & \textbf{LC WR (\%)} & \textbf{WR (\%)} & \textbf{Length} & \textbf{LC WR (\%)} & \textbf{WR (\%)} & \textbf{Length}  \\ \midrule
        \multirow{6}{*}{\tabincell{c}{\textbf{Llama-3-8b}\\ \textbf{-Instruct}}} & Initial & 22.92 & 22.57 & 1899 & 22.92 & 22.57 & 1899  \\ 
        ~ & DPO & 48.14 & \underline{51.52} & 2168 & 27.91 & 27.45 & 1945  \\ 
        ~ & SimPO & 47.56 & 40.72 & 1756 & 26.77 & 27.25 & 1984 \\ 
        ~ & SamPO & \underline{52.17} & 46.31 & 1806 & 26.95 & 27.16 & 1991  \\ 
        ~ & LDDPO & 52.1 & \textbf{51.72} & 2036 & \underline{28.55} & \underline{28.54} & 1956  \\ 
        ~ & OTPO & \textbf{53.37}\textsuperscript{***} & 47.58 & 1791 & \textbf{29.64}\textsuperscript{***} & \textbf{29.54} & 1991  \\ 
        \midrule
        \multirow{6}{*}{\tabincell{c}{\textbf{Llama-3.2-3b} \\ \textbf{-Instruct}}} & Initial & 17.97 & 19.34 & 2041 & 17.97 & 19.34 & 2041  \\ 
        ~ & DPO & 26.02 & \underline{27.96} & 2094 & 19.99 & \underline{20.54} & 1970  \\ 
        ~ & SimPO & 22.58 & 22.34 & 1944 & 19.03 & 19.72 & 1996  \\ 
        ~ & SamPO & 24.08 & 26.49 & 2084 & 18.58 & 19.65 & 2018  \\ 
        ~ & LDDPO & \underline{26.56} & 25.79 & 1909 & \underline{20.29} & 20.2 & 1939  \\ 
        ~ & OTPO & \textbf{26.97}\textsuperscript{***} & \textbf{28.61} & 2075 & \textbf{20.5}\textsuperscript{***} & \textbf{21.25} & 2000 \\
        
        \bottomrule
    \end{tabular}
    \caption{AlpacaEval 2 evaluation results under four settings. WR denotes win rate, LC WR denotes length-controlled win rate. Models aligned using OTPO achieve superior performance on length-controlled win rates across all settings. The best results are marked in \textbf{bold}. The second-best results are \underline{underlined}. Results marked with \textsuperscript{***} are significantly better than others with 99\% confidence.}
    \label{tab:main_results}
    \vspace{-10pt}
\end{table*}


In this section, we present the main results of our experiments, demonstrating the superior performance of OTPO in various settings (Sec.~\ref{sec:main_results}). Then we conduct ablation studies to validate the components of OTPO (Sec.~\ref{sec:ablation_study}). Finally, we conduct a human evaluation for thorough evaluation (Sec.~\ref{sec:human_evaluation}).

\subsection{Main Results}
\label{sec:main_results}

\begin{figure}[t]
    \centering
    \includegraphics[width=0.9\linewidth]{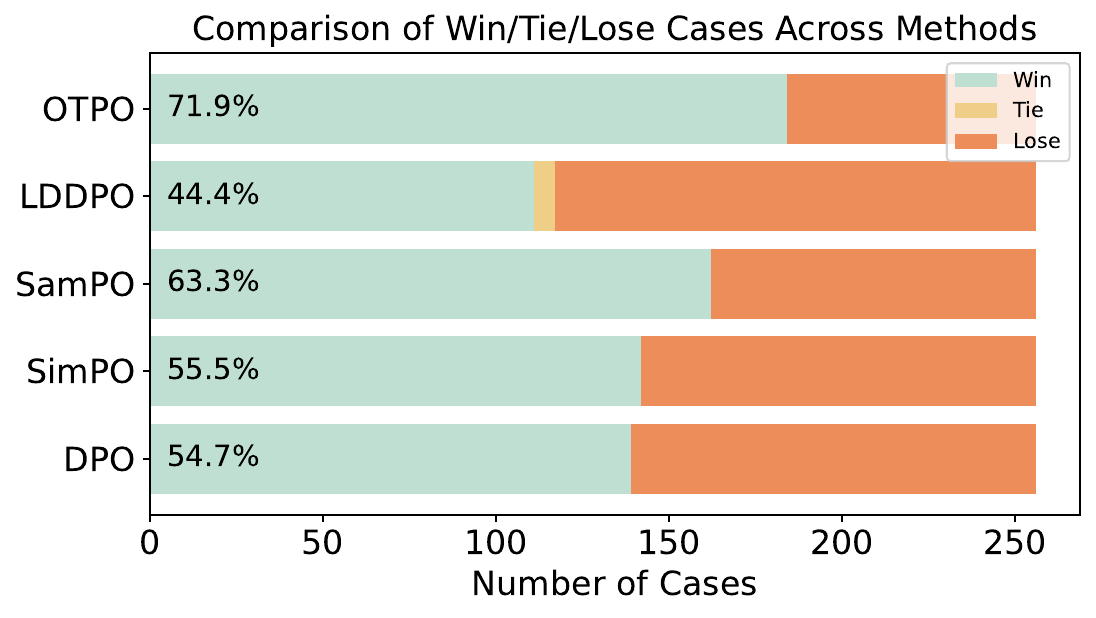}
    \caption{TL;DR summarization win rates compared to the base model, using GPT-4o as the evaluator. OTPO exceeds the existing methods by a large margin.}
    \label{fig:tldr_winrate}
    \vspace{-15pt}
\end{figure}

We present the main results in Tab.~\ref{tab:main_results} and Fig~\ref{fig:tldr_winrate},  showcasing OTPO's superiority in optimizing preferences across different backbones, tasks, and optimization strategies.

\textbf{Overall preference enhancement of OTPO.}
As shown in Tab.~\ref{tab:main_results}, OTPO demonstrates the best result on length-controlled win rate across 2 backbones and 2 datasets, with an improvement of 2.6\% to 10.9\% increase compared to DPO, and 1.0\% to 3.8\% increase compared other baselines.
Furthermore, it achieves the best win rate in 3 out of 4 settings, with up to 3.5\% increase compared to the best baseline, demonstrating its robust effectiveness in improving alignment. Notably, OTPO is also better than other methods as shown in Tab.~\ref{tab:compare_other_variants} in Appx.~\ref{app:evaluation_details}.

\textbf{OTPO excels in the domain-specific task.} In summarization task, OTPO exceeds other methods by a large margin of 8.6\% win rate compared to the best-performing baseline SamPO, as shown in Fig.~\ref{fig:tldr_winrate}. As OTPO is trained to emphasize important parts in a response, it generates summarizations that are more concise and include key meanings.
As the summarization task specifically requires concise summarizations, we regard all responses exceeding a certain length as ``Lose'' during evaluation following~\cite{stiennon2020learning}.

\textbf{Mitigating Length bias.}
Although OTPO appears relatively less performant in the UltraFeedback setting with Llama-3-8B based on the naive win rate of 47.58\%, this can be largely attributed to its production of substantially shorter responses, with an average length of 1791 tokens compared to the longer responses generated by other methods. 
Longer responses are not inherently problematic, especially if increased length leads to improved quality. However, \cite{zheng2023judging} has shown that ``LLM judge favors longer, verbose responses, even if they are not as clear, high-quality, or accurate as shorter alternatives.'' Our goal with OTPO is not to shorten responses arbitrarily but to prioritize essential, high-quality content, with brevity emerging as a natural byproduct. By mitigating unnecessary verbosity, OTPO ensures that models focus on delivering information more concisely without sacrificing informativeness, as indicated by the highest length-controlled win rate 53.37\%.

\textbf{Length bias varies across settings.}
The results reveal opposing trends in length bias between the two datasets, particularly with Llama-3.2-3b-Instruct. In the UltraFeedback setting, there is a 3\% increase in response length compared to the initial model, while in the HelpSteer2 setting, there is a 3\% decrease in response length. This suggests that length bias can manifest differently depending on the combination of the dataset and the optimized model, leading to overly lengthy or overly concise responses. We describe a more detailed analysis of generated on-policy datasets in Appx.~\ref{app:dataset_statistic}.

\subsection{Ablation Study}
\label{sec:ablation_study}


\begin{table}[t]
    \centering
    \begin{tabular}{cccc}
        \toprule
         & \textbf{LC WR} & \textbf{WR} & \textbf{Length} \\ 
        \midrule
        Initial & 22.92 & 22.57 & 1899 \\ 
        DPO & 48.14 & \underline{51.52} & 2168 \\ 
        OTPO & \textbf{53.37} & 47.58 & 1791 \\ 
        \midrule
        \multicolumn{4}{c}{\textit{(1) Ablation of Optimal Transport}} \\
        Uniform  & 52.60 & 46.36 & 1796 \\
        Similarity  & \underline{53.28} & 46.09 & 1757 \\
        \midrule
        \multicolumn{4}{c}{\textit{(2) Ablation of Weight Normalization}}\\
        None & 26.38 & 26.07 & 1939 \\
        Mean & 52.79 & 46.69 & 1791 \\ 
        Max & 49.85 & 44.77 & 1808 \\ 
        Length & 48.51 & \textbf{52.12} & 2167 \\ 
        \bottomrule
    \end{tabular}
    \caption{Ablation study of OTPO on Llama-3-8B-Instruct with UltraFeedback. We ablate each component of OTPO: (1) (middle part) Replace OT weight with uniform weight or cosine similarity-based weight. (2) (lower part) Varying the  weight sum normalization term $\tau$ by mean/max of the two responses' length, or each response length in Eq.~\ref{eq:rescale} after OT. }
    \label{tab:ablation_study}
    \vspace{-15pt}
\end{table}

We ablate each key design in OTPO: optimal transport guided weighting scheme, and weight normalization, and then report the results in Tab.~\ref{tab:ablation_study}.

\subsubsection{Alternatives of Token Weighting Scheme}

We replace the optimal transport-based token weighting scheme with uniform weight, or embedding similarity-based weight for comparison.
Uniform weight only ensures response length fairness, while embedding similarity-based weight additionally applies a simplified algorithm with a similar motivation as OTPO.

\textbf{Uniform weight ensures response length fairness}.
Uniform weight simply down-weight tokens in the longer response's weight to $\frac{\min(|y_c|, |y_r|)}{\max(|y_c|, |y_r|)}$ in DPO without OT (``Uniform''), so that each response's weight sum up to $\min(|y_c|, |y_r|)$. Compared to DPO, it decreases average response length (-372), thus trading off win rate (-10\%) against length-controlled win rate (+9.3\%).

\textbf{Similarity-based weighting scheme improves performance.}
The embedding similarity-based weighting method (``Similarity'') simplifies the OT process while adhering to the same intuition. For each token in the response, its representation is compared to the average representation of all tokens in the other response using cosine similarity. These similarities are then passed through a softmax function to calculate relative importance across all tokens in the response. Finally, we assign the same total weight budget $\min(|y_c|, |y_r|)$ based on the relative importance to obtain the final weight. 
This approach further improves length fairness by considering token-level relationships between the chosen and rejected responses.
However, it is slightly worse than OTPO in both length-controlled win rate and win rate, as it only considers the relationship of each token to the other response, while failing to consider more detailed token pair relationships.

\subsubsection{Impact of Weight Normalization}

We test various normalization strategies alongside OT, experimental indicates that they are worse than the original ``min'' normalization.

\begin{figure}[t]
    \centering
    \includegraphics[width=0.99\linewidth]{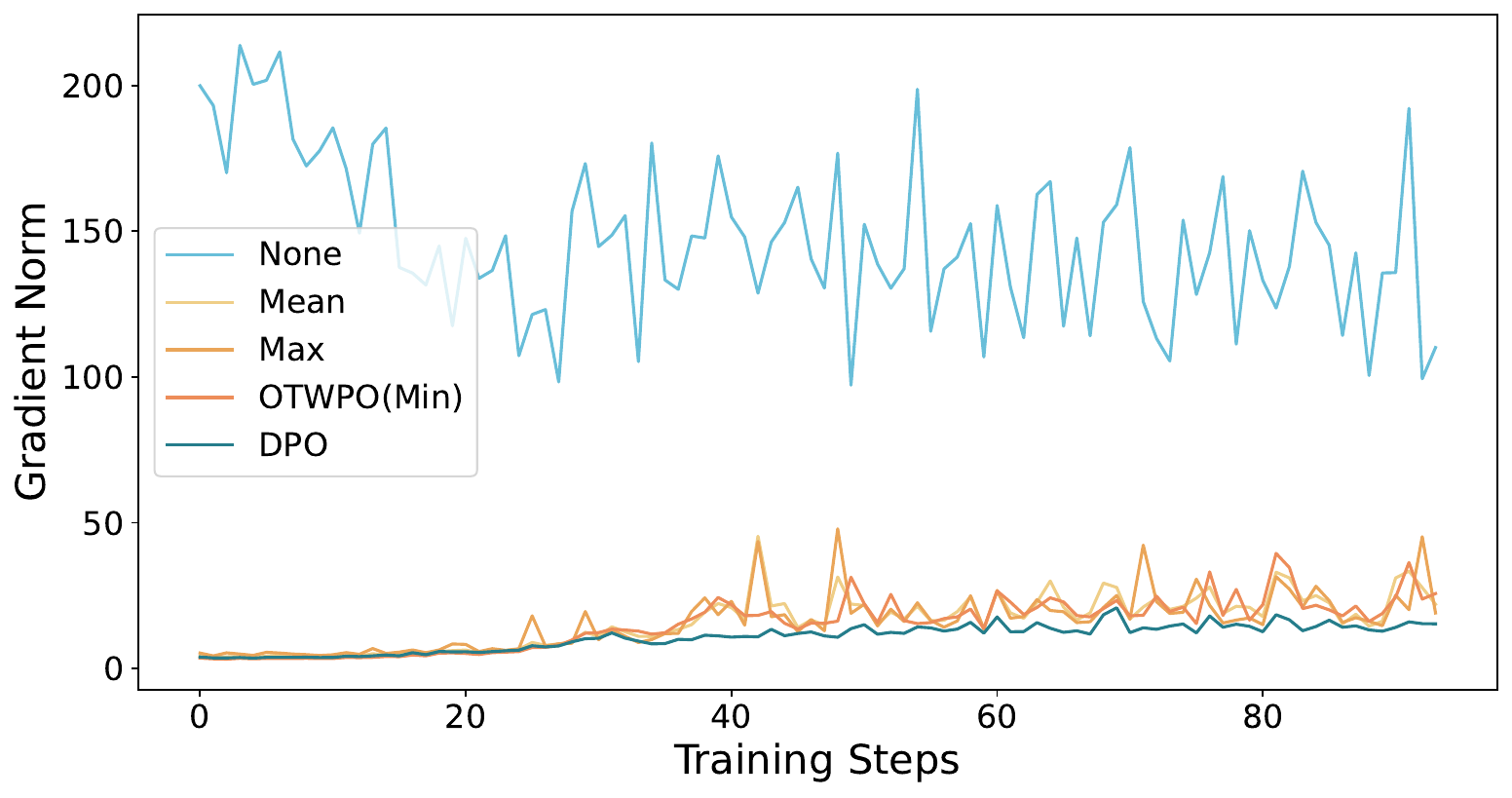}
    \caption{Trend of gradient norm during training.}
    \label{fig:grad_norm}
    \vspace{-15pt}
\end{figure}

\textbf{Other normalization variants lead to large fluctuations in training.}. We further report the trend of gradient norm during training in Fig.~\ref{fig:grad_norm}. Using no normalization (``None'') leads to significantly lower alignment performance, less than half of the best length-controlled win rate. This is mainly due to the large fluctuations in total weight $|\Gamma|$, thus leading to too aggressive gradient updates. Using mean or max for normalization leads to a decrease in both length-controlled win rate and win rate. This may relate to the large fluctuations in the shorter response's weight upscale across samples, which can lead to suboptimal performance.

\textbf{Separating normalization for each response achieves the best win rate.} We train a version where each response's weights are rescaled to match its original sum, i.e. $\tau=|y_c|$ for $y_c$ and $\tau=|y_r|$ for $y_r$, alongside the application of OT (``Length''). This approach, while slightly improving both the length-controlled win rate and win rate over the DPO baseline, lags behind OTPO in terms of length-controlled win rate as it leads to a mismatch between the chosen and rejected responses' distributions, which affects overall alignment. These findings demonstrate the complementary roles of OT and weight normalization in optimizing both response quality and alignment.

\subsection{Human Evaluation}
\label{sec:human_evaluation}

\begin{table}[t]
    \centering
    \begin{tabular}{ccc}
        \toprule
        ~ & \textbf{Expert1} & \textbf{Expert2}  \\ 
        \midrule
        \textbf{DPO} & 0.46 & 0.5   \\ 
        \textbf{SimPO} & \underline{0.56} & \underline{0.54} \\ 
        \textbf{SamPO} & 0.48 & 0.46 \\ 
        \textbf{LDDPO} & \underline{0.56} & 0.48 \\ 
        \textbf{OTPO} & \textbf{0.62} & \textbf{0.64} \\ 
        \bottomrule
    \end{tabular}
    \caption{Human evaluation of the win rates of different methods compared to the base model. OTPO is considered the best by both experts.}
    \vspace{-15pt}
    \label{tab:human_evaluation}
\end{table}

We conduct human evaluations to further verify the effectiveness of OTPO. We randomly sample 50 questions across multiple domains as input, prompt the base model, i.e. \texttt{Llama-3-8B-Instruct}, and the model aligned using UltraFeedback to answer the question. Then we ask human experts to choose the better response based on relevance, coherence, completeness, and conciseness. The two responses' positions are randomly swapped to ensure evaluation fairness. See more details of human evaluation's setting in Appx.~\ref{sec:human_evaluation_detail}. Tab.~\ref{tab:human_evaluation} reports win rates judged by human experts. 
We find the two experts' annotation results have a relatively low correlation of 0.37, which can be attributed to the diverse nature of human preference and the similarity across responses from the same model family. Results show that OTPO is considered the best among the two experts despite their diverse preferences.

\section{Complexity and Efficiency Analysis}

In this section, we analyze the computational and memory complexity of OTPO, mostly attributed to the optimal transport learning schema, and compare its efficiency with related methods.

\textbf{Time Complexity.} The optimal transport learning schema has a time complexity of $O(n^2)$, which is negligible compared to the transformer's forward pass complexity of $O(ln^2d + lnd^2)$, where $n$ is the input length, $d$ is the hidden dimension, and $l$  is the model depth.

\textbf{Memory Complexity.} The OT step requires storing the pairwise cost matrix $M \in \mathbb{R}^{l \times l}$, two auxiliary vectors in $\mathbb{R}^l$, and an additional matrix of the same shape as $M$. This results in a memory complexity of $O(l^2)$, which is minor compared to the memory requirement of the transformer forward and backward passes, typically $O(l(n^2 + nd^2))$. As a result, OTPO introduces negligible additional memory overhead.

\begin{figure}
    \centering
    \includegraphics[width=0.99\linewidth]{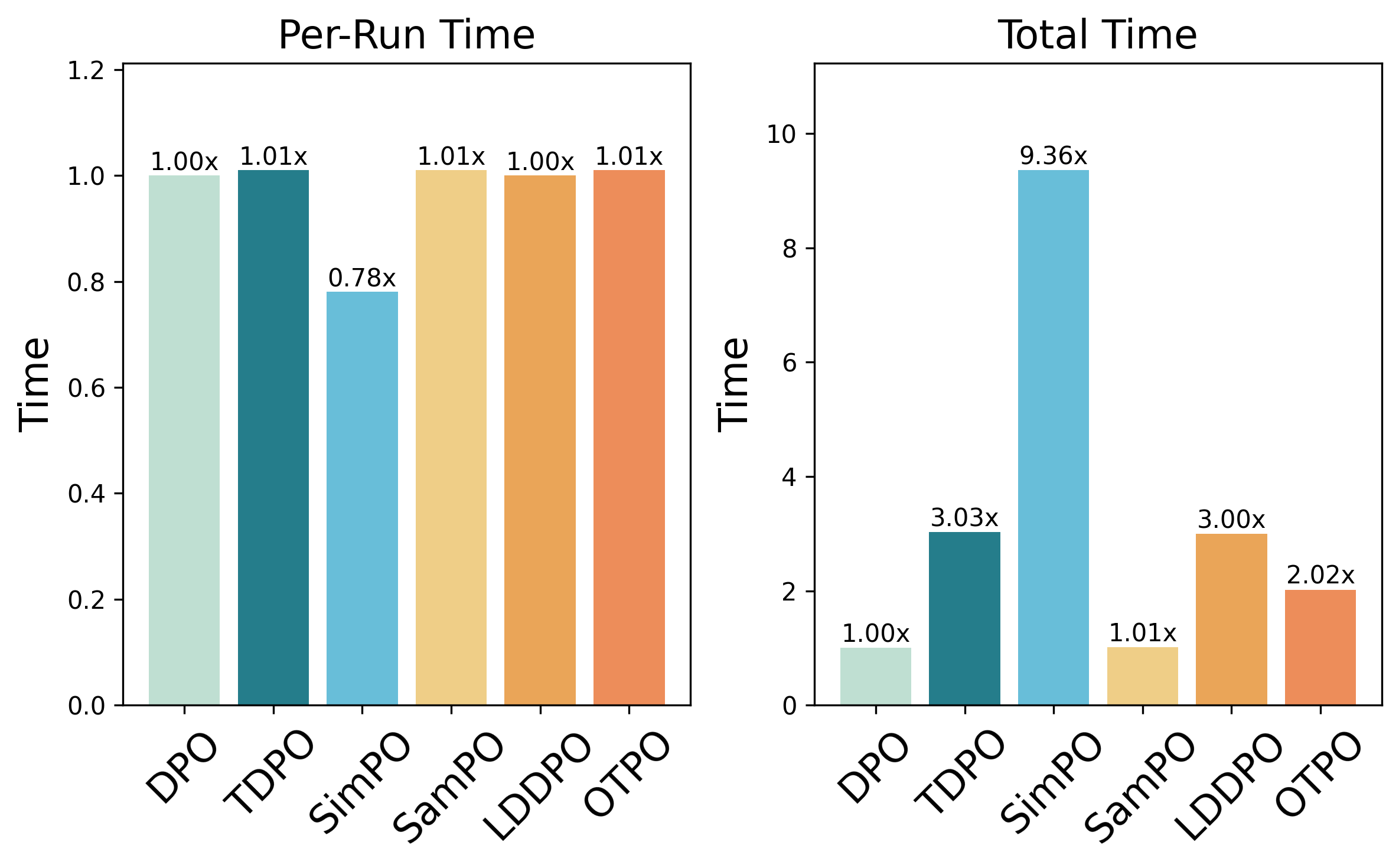}
    \caption{Comparison of training time across different preference optimization methods. We normalize DPO's training time as the baseline (1.00) and report each method's relative time as a fraction of it. \textit{Per-Run Time} is for a single training run; \textit{Total Time} includes time for hyperparameter tuning based on DPO’s optimal settings.}
    \label{fig:time_complexity}
    \vspace{-15pt}
\end{figure}

\textbf{Empirical Efficiency.} As shown in Fig.~\ref{fig:time_complexity}, OTPO exhibits training efficiency comparable to existing preference optimization methods. Despite the slight increase in per-run cost due to the OT computation, OTPO remains favorable in terms of total training time, especially when hyperparameter tuning is considered. Given the optimal $\beta$ for DPO, SimPO requires extensive tuning due to its sensitivity to its hyperparameters $\beta,\gamma$, while OTPO is more stable and thus requires fewer runs. 

In summary, OTPO introduces minimal overhead in both time and memory while offering improved instruction-following performance. This favorable trade-off makes OTPO a practical and efficient choice for alignment training.

\section{Related Work}

\textbf{Preference Optimization for LLMs.}
Preference optimization plays a vital role in aligning LLMs with human values and expectations. First proposed in~\cite{ouyang2022training}, they train the policy using proximal policy optimization to maximize the estimated reward given by a trained reward model. Direct preference optimization eliminates the need for a reward model and directly leverages pairwise comparisons to guide models toward preferred behaviors. Many efforts have been taken to improve the offline learning objective or to reduce computational costs. RSO~\cite{liu2023statistical}, IPO~\cite{azar2024general}, EXO~\cite{ji2024towards}, NCA~\cite{chen2024noise}, BCO~\cite{jung2024binary}, SPPO~\cite{wu2024self} replace the sigmoid function in DPO with other non-linear variants to model different preference objectives. CDPO~\cite{mitchell2023note}, Robust DPO (R-DPO)~\cite{ray2024provably} enhances DPO by improving robustness to preference noise, and  RPO~\cite{pang2024iterative} adds a negative log-likelihood term to prevent large decrease in chosen responses' probability. To tackle the commonly observed length bias, SimPO~\cite{meng2024simpo}, SamPO~\cite{lu2024eliminating}, LDDPO~\cite{liu2024length} apply a heuristical kind of token weighting scheme to elevate length fairness between two responses.
APO~\cite{d2024anchored} rewrites irrelevant parts with an external LLM to create minimally contrastive preference data.
Our work builds upon these insights, introducing optimal transport to calculate token weights by assigning more importance to those semantically relevant tokens.

\noindent\textbf{Optimal Transport for Machine Learning.} 
Optimal Transport has proven to be a powerful tool in machine learning, particularly for tasks involving distribution alignment, such as transfer learning~\cite{flamary2016optimal,courty2017joint}, generative modeling, e.g. Wasserstein GANs~\cite{arjovsky2017wasserstein} and \cite{wang2024autoencoder}, natural language processing~\cite{asano2020self}, and recommendation~\cite{han2024intra}. In particular, recent efforts have also utilized OT to perform preference optimization on unpaired preference datasets by achieving distributional dominance~\cite{melnyk2024distributional}. We apply OT to align token distributions between chosen and rejected responses in preference optimization, capturing the fine-grained differences in token semantics and context, and enabling a more principled weighting mechanism.

\section{Conclusion}

In this paper, we proposed an \textbf{O}ptimal \textbf{T}ransport-based token weighting scheme for direct \textbf{P}reference \textbf{O}ptimization (OTPO), a context-aware token weighting scheme to reinforce semantically meaningful differences in reward estimation. 
OTPO leverages optimal transport to dynamically assign a fixed total weight budget to each token pair in the chosen and rejected response based on their semantic similarity, and then aggregate each token pair's log-likelihood ratio difference as a contrastive reward difference estimate. It represents a step toward more robust and interpretable reward estimation and lays the groundwork for future exploration into fine-grained preference modeling in alignment tasks.

\section*{Limitations}
Our experiments were limited to off-policy and on-policy setups for direct preference optimization. Recent research has highlighted that iterative on-policy setups may yield larger improvements in instruction-following performance. We did not explore such setups due to limited computational resources. This leaves room for future work to further enhance model performance using OTPO under iterative on-policy setups.

We conducted our experiments on relatively small models. While this scale provides meaningful insights, the scalability and generalizability of our algorithms to larger models, such as those with hundreds of billions of parameters, remain to be validated. Addressing this limitation would require significant computational resources and could further confirm the robustness of our approach across different model sizes.

For evaluating alignment quality, we relied on GPT-4 as the evaluation judge. While GPT-4 offers state-of-the-art evaluation capabilities, it may introduce potential biases and result in less accurate or reliable judgments. This could affect the evaluation of alignment improvements, and future work may explore more robust, unbiased, and possibly human-in-the-loop evaluation mechanisms.

\section*{Ethical Statements}
We performed only a relatively simple check on the datasets used in our experiments. Although we made efforts to ensure the datasets were suitable for training alignment models, they may contain few harmful or inappropriate content. Addressing these issues requires more thorough dataset curation and filtering processes, which were beyond the scope of this work.

Due to the availability of datasets, our training experiments were conducted mainly on English datasets. We have not verified the generalizability or effectiveness of our algorithm on non-English datasets. This limitation underscores the need for future work to ensure alignment algorithms are robust across languages and culturally diverse contexts.

Our primary focus was on improving instruction-following abilities in aligned large language models. However, these models may still exhibit safety risks, such as generating harmful or biased outputs, which were not fully addressed in this study. Post-alignment safety evaluations and interventions are critical to mitigate such risks and ensure the responsible deployment of these models.

\section*{Acknowledgements}

The authors would like to thank members of Shopee LLM Team for their helpful feedback and discussions, and the anonymous reviewers for their valuable suggestions.
This work was partly supported by the National Natural Science Foundation of China (NSFC) (NO. 62476279, NO. U2436209), Major Innovation \& Planning Interdisciplinary Platform for the ``Double-First Class'' Initiative, Renmin University of China, the Fundamental Research Funds for the Central Universities, and the Research Funds of Renmin University of China No. 24XNKJ18. This work was partially supported by fund for building world-class universities (disciplines) of Renmin University of China and Public Computing Cloud, Renmin University of China.

\bibliography{ref}

\appendix

\section{Implementation Details}
\label{app:implementation_details}

We use UltraFeedback~\cite{cui2024ultrafeedback}, the preference-transformed HelpSteer2~\cite{wang2024helpsteer2}, and TL;DR~\cite{stiennon2020learning} as preference training datasets. 
UltraFeedback contains 61,135 examples in its training split, and HelpSteer2 contains 9125 examples. TL;DR contains 92,534 summary comparisons for training, and 83,629 test comparisons.
We generate the on-policy training data as follows: 1) Sample $n$ responses for each prompt with the policy to be optimized. 2) Discard those samples where the $n$ responses are identical. 3) Annotate the responses with a reward model, and select the response with the highest and lowest score as $y_c, y_r$.
We set $n=5$ for UltraFeedback, and $n=10$ for HelpSteer2 following~\cite{meng2024simpo,tunstall2023zephyr,wang2024helpsteer2}. The detailed statistics of the on-policy datasets are listed in Appx.~\ref{app:dataset_statistic}.
Maximum prompt length and maximum input length are controlled as (2048, 1800) for the instruction-following task and (1024, 900) for the summarization task. This setting accommodates varying input sizes and prevents huge memory costs.

We first conduct preliminary hyperparameter search on learning rate in $\{3e^{-7}, 5e^{-7}, 7e^{-7}\}$, batch size in $\{64, 128, 256\}$, and epoch in $\{1, 2, 3\}$. Results show that training with a learning rate of $5e^{-7}$ and a batch size of $128$ for 1 epoch typically yields the best results, so we use this set of hyperparameters for all experiments. Then we search $\beta$ in $\{0.01, 0.05, 0.1\}$ for each setting using DPO, and set $\beta=0.01$ for Llama-3-8B and Qwen-2.5-3B, $\beta=0.1$ for Llama-3.2-3B across methods except for SimPO, which requires a much larger $\beta$. As for method-specific hyperparameters, we report the search ranges considered in Tab.~\ref{tab:hyperparams_baseline}. Optimization was performed using AdamW with a cosine schedule and warmup ratio of 0.1.

\begin{table*}[!ht]

\centering
\resizebox{\textwidth}{!}{
    \begin{tabular}{lll}
        \toprule 
        \textbf{Method} & \textbf{Loss Function} & \textbf{Hyperparameter} \\ \midrule
        
        DPO & $-\log \sigma ( \beta \log \frac{\pi_\theta(y_c|x)}{\pi_{\text{ref}}(y_c|x)} - \beta \log \frac{\pi_\theta(y_r|x)}{\pi_{\text{ref}}(y_r|x)})$ & $\beta \in [0.01, 0.05, 0.1]$ \\  \bottomrule

        \multirow{2}{*}{SimPO} & \multirow{2}{*}{$-\log \sigma  ( \frac{\beta}{|y_c|} \log \pi_\theta(y_c|x) - \frac{\beta}{|y_r|} \log \pi_\theta(y_r|x) - \gamma )$} & $\beta \in [2.0, 2.5, 10, 20]$ \\
        & & $\gamma \in [1, 2, 3]$ \\
        \midrule
        
        \multirow{2}{*}{SamPO} & $-\log \sigma ( \beta \sum_{t=1}^{m}\log \frac{\pi_\theta(y_c^t|x)}{\pi_{\text{ref}}(y_c^t|x)} - \beta \sum_{t=1}^{m} \log \frac{\pi_\theta(y_r^m|x)}{\pi_{\text{ref}}(y_r^m|x)}), $   &  \multirow{2}{*}{-}  \\
        ~ & $ \text{where} \, m = min(|y_c|, |y_r|), y^t \sim \text{Uniform}(m, \{y\}^T)$ & ~ \\

        \midrule

        \multirow{2}{*}{LDDPO} & $-\log \sigma ( \beta (\sum_{t=1}^{m}\log \frac{\pi_\theta(y_c^t|x)}{\pi_{\text{ref}}(y_c^t|x)} + \alpha \sum_{t=m+1}^{|y_c|}\log \frac{\pi_\theta(y_c^t|x)}{\pi_{\text{ref}}(y_c^t|x)}) $ & \multirow{2}{*}{$\alpha \in [0.2, 0.5, 0.7]$} \\
        ~ & $  - \beta (\sum_{t=1}^{m} \log \frac{\pi_\theta(y_r^t|x)}{\pi_{\text{ref}}(y_r^t|x)} + \alpha \sum_{t=m+1}^{|y_r|} \log \frac{\pi_\theta(y_r^t|x)}{\pi_{\text{ref}}(y_r^t|x)} ) ), m = min(|y_c|, |y_r|) $  &  \\

        \midrule
        \multirow{3}{*}{TDPO} & $-\log \sigma((\beta \log \frac{\pi_{\theta}(y_c | x)}{\pi_{\text{ref}}(y_c | x)}-\beta \log \frac{\pi_{\theta}(y_r | x)}{\pi_{\text{ref}}(y_r | x)})$ & \multirow{3}{*}{$\alpha \in [0.1, 0.5, 1.0]$}\\
        
        ~ & $-\alpha(\beta D_{\mathrm{SeqKL}}(x, y_r ; \pi_{\mathrm{ref}} \| \pi_\theta)-\mathnormal{sg}(\beta D_{\mathrm{SeqKL}}(x, y_c ; \pi_{\mathrm{ref}} \| \pi_\theta))),$ & ~\\
        ~ & $ \text{where} \, D_{\mathrm{SeqKL}}(x, y;\pi_{\text{ref}}\|\pi_{\theta})=
        \sum_{t=1}^{|y|} D_{\mathrm{KL}}(\pi_{\text{ref}}(\cdot|[{x}, y^{<t}])\|\pi_{\theta}(\cdot|[{x}, y^{<t}]))$ & ~ \\

        \midrule
        OTPO & $-\log \sigma(\beta ( \sum_{i=1}^{|y_c|} \omega_c^{*i} \log \frac{\pi_\theta(y_c^i|x, y_c^{<i})}{\pi_{\text{ref}}(y_c^i|x, y_c^{<i})} 
        - \sum_{j=1}^{|y_r|} \omega_r^{*j} \log \frac{\pi_\theta(y_r^j|x, y_r^{<j})}{\pi_{\text{ref}}(y_r^j|x, y_r^{<j})}))$ & $\epsilon_1 \in [0.1, 1],\; \epsilon_2 =0.2$ \\
        
        \bottomrule
    \end{tabular}
}
\caption{Various preference optimization loss functions with weighting scheme and hyperparameters search range. Here, $\text{Uniform}(m, \{y\}^T)$ denotes uniformly sampling $m$ tokens from all tokens in $\{y\}^T$, and $\mathnormal{sg}$ represents the stop-gradient operator, which blocks the propagation of gradients.}
\label{tab:hyperparams_baseline}
\vspace{-10pt}
\end{table*}

For the Optimal Transport part, each token's last-layer hidden state was used as its representation. We set \(\epsilon_1=1\) for UltraFeedback and TL;DR, \(\epsilon_1=0.1\) for HelpSteer2, and \(\epsilon_2=0.2\) for all configurations. As for supervise fine-tuning Qwen-2.5-3B, we apply a batch size of 128 and a learning rate of $2e{-5}$ for 1 epoch.

We use the \href{https://github.com/huggingface/alignment-handbook}{\texttt{alignment-handbook}} library with Apache-2.0 license to perform preference optimization and supervise fine-tuning, incorporating the \href{https://pythonot.github.io/index.html}{\texttt{POT}}~\cite{flamary2021pot} package to compute the optimal transport plan. For evaluation, we assess the model's instruction-following capabilities using the official \href{https://github.com/tatsu-lab/alpaca_eval}{\texttt{AlpacaEval2}} repository. Additionally, we evaluate the model's general abilities using the \href{https://github.com/EleutherAI/lm-evaluation-harness/tree/main}{Harness} evaluation framework~\cite{eval-harness}.

As for the computation environment, we conducted training on 4xA100 for Llama-3-8B, and 2xA100 for Llama-3.2-3B, Qwen-2.5-3B using Pytorch. To accelerate training, we utilized FlashAttention2~\cite{dao2024flashattention}, DeepSpeed Zero 3~\cite{rasley2020deepspeed,ren2021zero}, and bfloat16 precision. 

\section{On-policy Dataset Statistics}
\label{app:dataset_statistic}

\begin{table}
    \centering
    \begin{tabular}{ccccc}
        \toprule
        ~ & \multicolumn{2}{c}{\textbf{Llama-3-8B}} & \multicolumn{2}{c}{\textbf{Llama-3.2-3B}} \\
        \cmidrule(lr){2-3} \cmidrule(lr){4-5}
        ~ & UF & HS & UF & HS \\ 
        \midrule
        \textbf{count} & 59876  & 9084  & 60692  & 9087  \\ 
        \textbf{mean} & -29 & -24 & -91 & -75 \\ 
        \textbf{std} & 233 & 168 & 447 & 257 \\ 
        \textbf{min} & -4058  & -1854  & -4094  & -2016  \\ 
        \textbf{25\%} & -67  & -75  & -99  & -115  \\ 
        \textbf{50\%} & -11  & -14  & -16  & -30  \\ 
        \textbf{75\%} & 31  & 37  & 30  & 28  \\ 
        \textbf{max} & 2218  & 1193  & 4079  & 1266  \\ 
        \bottomrule
    \end{tabular}
    \caption{Response length difference summary statistic of each on-policy dataset. UF denotes UltraFeedBack, while HS denotes HelpSteer2.}
    \label{tab:response_length_stat}
\end{table}

\begin{figure}[t]
    \centering
    \includegraphics[width=0.99\linewidth]{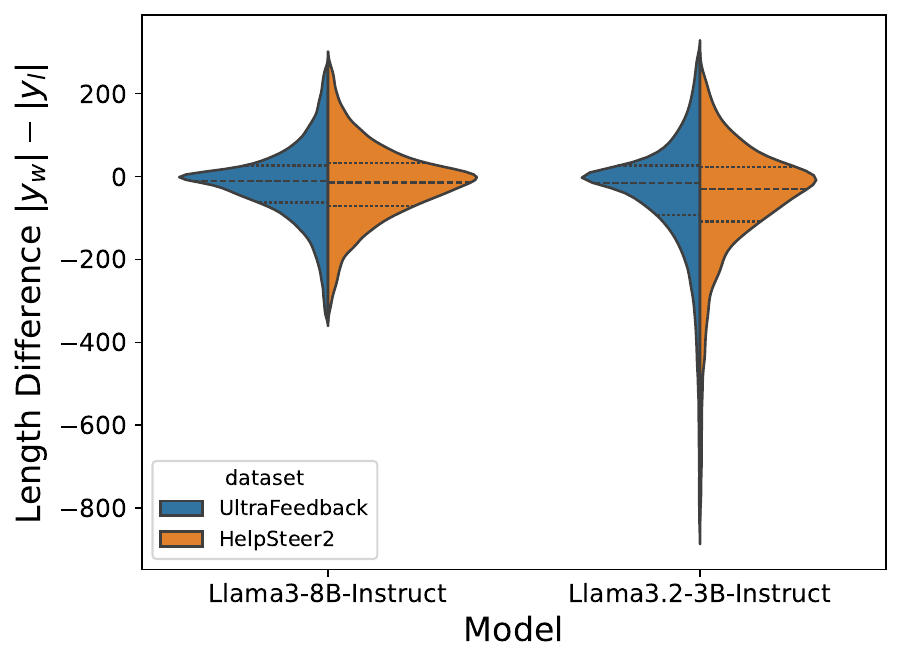}
    \caption{Dataset length difference ($|y_c| - |y_r|$) distribution. Dash lines indicated the quartiles.}
    \label{fig:dataset_length_distribution}
\end{figure}

We present the statistics of length differences in the generated on-policy datasets in Tab.~\ref{tab:response_length_stat}. To provide a clearer visualization of the length difference distribution, we leave out the top and bottom 2.5\% of samples and visualize the remaining data using a violin plot in Fig.~\ref{fig:dataset_length_distribution}. The results show that prompts in HelpSteer2 tend to produce preference data pairs with more negative length differences, having a median of -30 compared to -16 for UltraFeedback with Llama-3.2-8B. Notably, the length differences in the long-tail regions are significantly smaller for Llama-3.2-3B, further highlighting the contrast between the two models.

We sample 50 samples from each on-policy dataset to verify that the data does not contain any information that uniquely identifies individual people of offensive content.

\section{General Ability Evaluation}
\label{app:evaluation_details}

\begin{table*}[p]
    \centering
    \begin{tabular}{ccccccc}
        \toprule
        ~ & \textbf{MMLU} & \textbf{GSM8K} & \textbf{ARC} & \textbf{HellaSwag} & \textbf{PiQA} & \textbf{Average} \\
        
        \midrule
        \multicolumn{7}{c}{\textit{Llama-3-8b-Instruct + UltraFeedback}} \\ 
        \midrule
        \textbf{Initial} & 65.64  & 75.59  & 56.57  & 57.70  & 78.18  & 66.74  \\ 
        \textbf{DPO} & 65.80  & 74.98  & 56.83  & 56.07  & 74.92  & 65.72  \\ 
        \textbf{SimPO} & 65.75  & 75.13  & 56.06  & 54.41  & 75.73  & 65.42  \\ 
        \textbf{SamPO} & 65.62  & 70.43  & 55.46  & 53.98  & 74.21  & 63.94  \\ 
        \textbf{LDDPO} & 65.73  & 71.65  & 57.17  & 55.13  & 74.70  & 64.87  \\ 
        \textbf{OTPO} & 65.48  & 70.51  & 55.29  & 53.94  & 74.48  & 63.94  \\ 
        
        \midrule
        \multicolumn{7}{c}{\textit{Llama-3-8b-Instruct + HelpSteer2}} \\
        \midrule
        \textbf{Initial} & 65.64  & 75.59  & 56.57  & 57.70  & 78.18  & 66.74  \\ 
        \textbf{DPO} & 65.79  & 75.97  & 56.66  & 57.87  & 78.73  & 67.00  \\ 
        \textbf{SimPO} & 65.72  & 75.36  & 56.66  & 57.79  & 78.40  & 66.79  \\ 
        \textbf{SamPO} & 65.81  & 75.66  & 56.91  & 57.91  & 78.40  & 66.94  \\ 
        \textbf{LDDPO} & 65.69  & 75.13  & 56.74  & 57.93  & 78.40  & 66.78  \\ 
        \textbf{OTPO} & 65.80  & 76.04  & 57.17  & 57.88  & 78.35  & 67.05  \\ 
        
        \midrule
        \multicolumn{7}{c}{\textit{Llama-3.2-3b-Instruct + UltraFeedback}} \\ 
        \midrule
        \textbf{Initial} & 59.71  & 64.14  & 45.90  & 52.34  & 75.63  & 59.54  \\ 
        \textbf{DPO} & 60.06  & 66.49  & 46.84  & 52.48  & 75.84  & 60.34  \\ 
        \textbf{SimPO} & 59.84 & 64.97 & 47.18 & 52.54 & 75.73 & 60.05 \\ 
        \textbf{SamPO} & 59.72  & 65.88  & 46.84  & 52.53  & 76.01  & 60.20  \\ 
        \textbf{LDDPO} & 60.00 & 66.03 & 47.18 & 52.49 & 75.79 & 60.30 \\ 
        \textbf{OTPO} & 59.98  & 66.41  & 46.67  & 52.50  & 75.90  & 60.29  \\ 
        
        \midrule
        \multicolumn{7}{c}{\textit{Llama-3.2-3b-Instruct + HelpSteer2}} \\ 
        \midrule
        \textbf{Initial} & 59.71  & 64.14  & 45.90  & 52.34  & 75.63  & 59.54  \\ 
        \textbf{DPO} & 59.64  & 64.29  & 46.16  & 52.38  & 75.68  & 59.63  \\ 
        \textbf{SimPO} & 59.78  & 64.22  & 46.16  & 52.33  & 75.84  & 59.67  \\ 
        \textbf{SamPO} & 59.68  & 63.53  & 46.08  & 52.22  & 75.90  & 59.48  \\ 
        \textbf{LDDPO} & 59.67  & 63.99  & 46.16  & 52.37  & 75.63  & 59.56  \\ 
        \textbf{OTPO} & 59.71  & 64.44  & 46.33  & 52.31  & 75.84  & 59.73  \\ 
        \bottomrule
    \end{tabular}

    \caption{General ability evaluation results across 4 settings.}
    \label{tab:general_ability}
\end{table*} 

\subsection{Benchmark Description}
To further validate the effect of OTPO on the models' general abilities, we evaluate the aligned models using 5 popular benchmarks:
\begin{itemize}
    \item MMLU~\cite{hendrycks2021measuring}: A massive multitask language understanding benchmark spanning broad domains. It consists of 4 subgroups and 57 subsets. We select the model's answer based on the probabilities of `A', `B', `C', and `D', as suggested in the original paper, and report the overall accuracy.
    \item GSM8K~\cite{cobbe2021training}: A mathematical benchmark of grade school math problems for evaluating reasoning abilities. We evaluate the original test split with 1.32k examples and report the accuracy of answers with strict exact match.
    \item ARC Challenge~\cite{clark2018think}: A more challenging benchmark involving complex reasoning over a diverse set of science exam questions, containing 2590 examples in the format of multiple choice. We report the normalized accuracy of overall test samples.
    \item HellaSwag~\cite{zellers2019hellaswag}: A benchmark for predicting the endings of stories or scenarios, evaluating LLM's comprehension and creativity, targeted at commonsense reasoning via natural language. We evaluate the test split containing 10k examples and report the overall accuracy.
    \item PiQA~\cite{bisk2020piqa}: A Physical Interaction Question Answering task to test physical commonsense reasoning, i.e. interaction with everyday objects in everyday situations. It contains 20k QA pairs that are either multiple-choice or true/false questions. We report the overall accuracy.
\end{itemize}

\subsection{Evaluation Results}

We report the general ability evaluation reports across 4 settings in Tab.~\ref{tab:general_ability}

\textbf{General ability assessment.}
OTPO exhibits comparable general ability to other baselines, with fluctuations less than 1\% in most of the setups. 
It consistently achieves the highest average scores across both Llama-3-8B-Instruct (67.05) and Llama-3.2-3b-Instruct (59.73) in the HelpSteer2 setup. Moreover, it excels in reasoning ability, resulting in an increase in GSM8K (+0.15\%) and ARC (+0.26\%) compared to the best baseline. This highlights OTPO's potential to capture contextual and semantic differences to improve the model's reasoning ability. 

\textbf{Training dataset comparison.}
Helpsteer2 ensures consistent general ability improvement across most of the methods. This can be attributed to the difference in data distribution. While UltraFeedback mostly contains open-ended questions, HelpSteer2 contains more reasoning-related questions. However, training OTPO with UltraFeedback leads to a drastic decrease in Llama-3-8B-Instruct (-2.8), specifically on  GSM8K(-5.08) and PiQA (-3.7). 
This phenomenon is commonly referred to as ``alignment-tax'', where models trade-off between general ability and instruction-following ability.
This calls for more rigorous consideration when choosing the dataset for alignment according to specific alignment needs.

\section{Experiments on Other Models}

We conducted additional experiments on Qwen-2.5-3B-Instruct and Mistral-7B-Instruct-v0.2 with on-policy datasets created from HelpSteer2 to further validate OTPO's effectiveness. The results are shown in Tab.~\ref{tab:more_backbones}

\begin{table*}[!ht]
    \centering
    \begin{tabular}{ccccccc}
        \toprule
        \multirow{2}{*}{\textbf{Method}} &  \multicolumn{3}{c}{\textbf{Qwen-2.5-3B-Instruct}} &  \multicolumn{3}{c}{\textbf{Mistral-7B-Instruct-v0.2}} \\ 
        \cmidrule(lr){2-4} \cmidrule(lr){5-7}
        ~ & LC WR & WR & length & LC WR & WR & length \\ 
        \midrule
        \textbf{Initial} & 16.82 & 20.1 & 2145 & 17.74 & 15.05 & 1630 \\ 
        \textbf{DPO} & 16.93 & 19.62 & 2100 & 27.25 & 24.78 & 1775 \\ 
        \textbf{SimPO} & 16.75 & 19.47 & 2112 & 23.56 & 22 & 1906 \\ 
        \textbf{SamPO} & \underline{18.26} & \underline{20.75} & 2105 & 27.31 & 25.31 & 1844 \\ 
        \textbf{LDDPO} & 17.32 & 20.15 & 2107 & \underline{28.22} & \underline{25.84} & 1815 \\ 
        \textbf{OTPO} & \textbf{19.71}\textsuperscript{***} & \textbf{22.06} & 2074 & \textbf{30.35}\textsuperscript{***} & \textbf{28.2} & 1839 \\ 
        \bottomrule
    \end{tabular}
    \caption{AlpacaEval 2 evaluation results with HelpSteer2 as the training dataset, Qwen-2.5-3B-Instruct and Mistral-7B-Instruct-v0.2 as the backbone model. WR denotes win rate, LC WR denotes length-controlled win rate. Models aligned using OTPO achieve the best performance on length-controlled win rates and win rates on both models. The best results are marked in \textbf{bold}. The second-best results are \underline{underlined}. Results marked with \textsuperscript{***} are significantly better than others with 99\% confidence.}
    \label{tab:more_backbones}
\end{table*}

\section{Comparison to Other DPO Variants}

\begin{table}[t]
    \centering
    \begin{tabular}{cccc}
        \toprule
        ~ & \textbf{LC WR (\%)} & \textbf{WR (\%)} & \textbf{Length} \\ 
        \midrule
        Initial & 22.92 & 22.57 & 1899 \\ 
        DPO & 27.91 & 27.45 & 1945 \\ 
        AOT & 28.55 & 27.83 & 1924 \\ 
        BCO & \underline{29.23} & 28.76 & 1955 \\ 
        CDPO & 28.62 & 28.52 & 1954 \\ 
        EXO & 27.22 & 26.82 & 1937 \\
        IPO & 25.02 & 26.06 & 2021 \\
        NCA & 27.72 & 27.82 & 1951 \\
        R-DPO & 28.66 & 28.34 & 1940 \\ 
        RSO & 28.93 & \underline{28.8} & 1961 \\ 
        RPO & 27.43 & 27.41 & 1963 \\
        SPPO & 28.48 & 28.13 & 1946 \\ 
        LR-DPO & 28.23 & 22.53 & 1654 \\
        \midrule
        SimPO & 26.77 & 27.25 & 1984 \\ 
        SamPO & 26.95 & 27.16 & 1991 \\ 
        LDDPO & 28.55 & 28.54 & 1956 \\ 
        OTPO & \textbf{29.64} & \textbf{29.54} & 1991 \\ 
        \bottomrule
    \end{tabular}
    \caption{A more comprehensive comparison of DPO variants on AlpacaEval2, with the lower part incorporating an implicit token weighting scheme. The best results are marked in \textbf{bold}. The second-best results are \underline{underlined}.}
    \label{tab:compare_other_variants}
\end{table}

OTPO remains the best algorithm compared to other DPO variants, along with the ones incorporating a certain kind of weighting scheme. We train Llama-3-8B on the HelpSteer2 dataset using these kinds of preference optimization loss:
\begin{itemize}
    \item AOT~\cite{melnyk2024distributional} align LLMs by making the reward distribution of the chosen responses stochastically dominant in the first order on the distribution of rejected samples.
    \item BCO~\cite{jung2024binary} trains a binary classifier that maps the chosen response to 1 and the rejected response to 0.
    \item CDPO~\cite{mitchell2023note} applies label smoothing to the original DPO loss, enhancing robustness to preference label noise.
    \item EXO~\cite{ji2024towards} replaces the forward KL with reverse KL when deriving DPO loss.
    \item IPO~\cite{azar2024general} minimizes the squared loss of margin between the estimated reward margin and a predefined margin.
    \item NCA~\cite{chen2024noise} optimizes the absolute likelihood of each response instead of the relative likelihood of two responses.
    \item R-DPO~\cite{ray2024provably} model the probability of existing label noise and apply label smoothing.
    \item RSO~\cite{liu2023statistical} replaces the sigmoid function with hinge loss.
    \item RPO~\cite{pang2024iterative} adds a negative log-likelihood loss of the chosen response to DPO loss, alleviating the decrease in the chosen reward.
    \item SPPO~\cite{wu2024self} treats the chosen and rejected response as two players, and solves the Nash equilibrium by pushing the chosen reward to 1/2 and the rejected reward to -1/2
    \item LR-DPO~\cite{park2024disentangling} adds a length regularization term by adding a weighted length difference to the reward difference term.
\end{itemize}
We leave the incorporation of an optimal transport-based weighting scheme to these DPO variants for future work.

\begin{figure}[t]
	\centering
	\begin{minipage}{0.99\linewidth}
		\centering
		\includegraphics[width=0.99\linewidth]{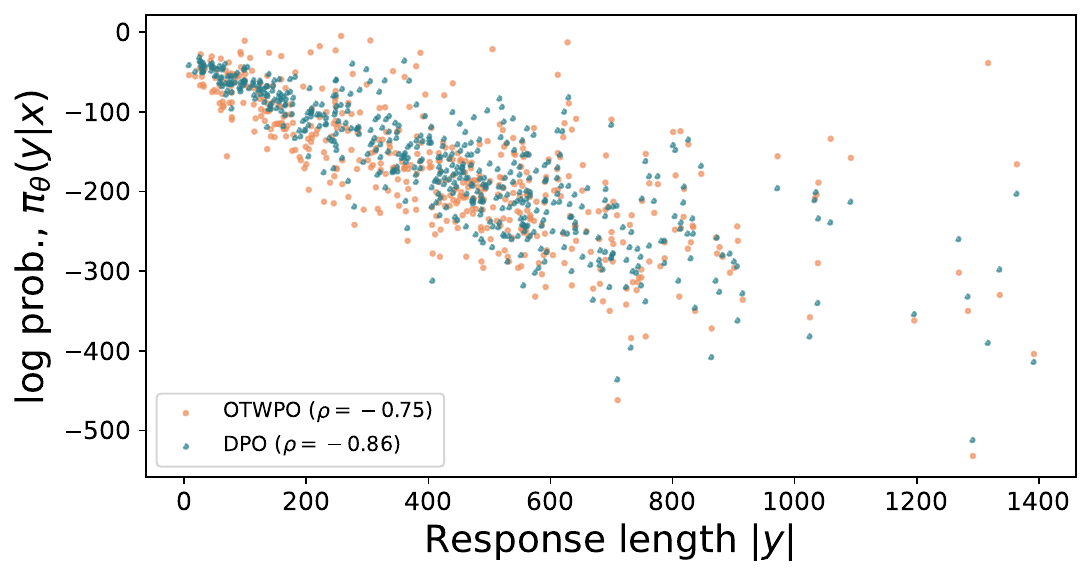}
		\caption{Estimated response log probability v.s. response length. OTPO exhibits a less positive trend in log probability with respect to DPO, leading to a more stable reward estimate.}
		\label{fig:logp_length}
	\end{minipage}
        \qquad
	
	\begin{minipage}{0.99\linewidth}
		\centering
		\includegraphics[width=0.90\linewidth]{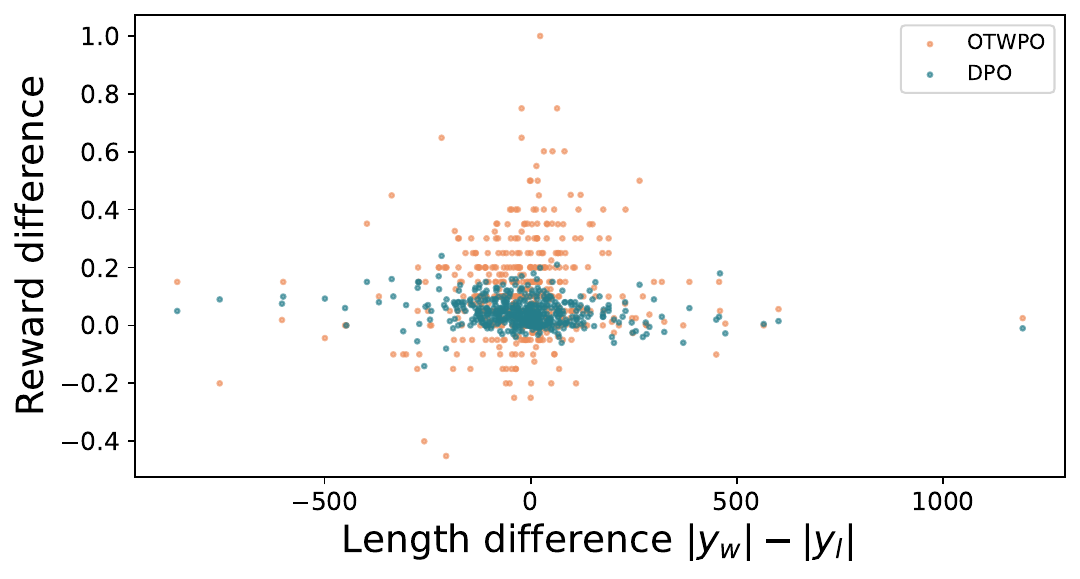}
		\caption{Reward difference v.s. the length difference between chosen and rejected responses. While the optimized reward differences by DPO are concentrated around 0.1, the reward differences optimized by OWTPO are more varied, especially larger when the length difference is low. }
		\label{fig:reward_length_diff}
	\end{minipage}
    \vspace{-15pt}
\end{figure}

\section{Hyperparameter Sensitivity}
\label{sec:hyperparameter_sensitivity}

\begin{figure*}[!t]
    \centering
    \includegraphics[width=0.99\linewidth]{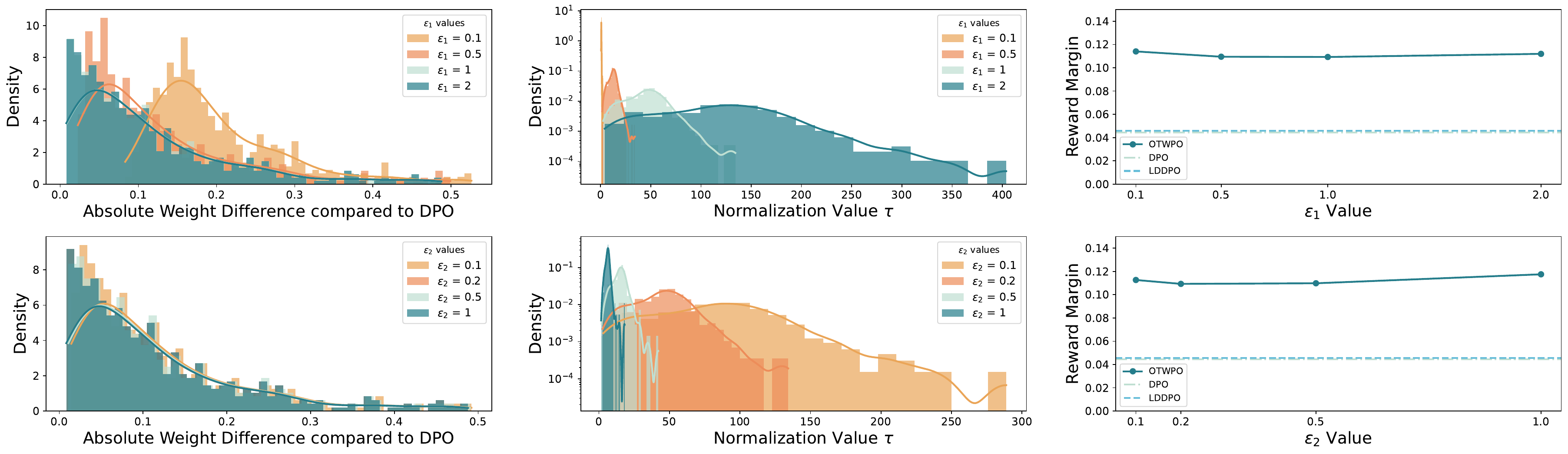}
\caption{Hyperparameter sensitivity analysis of $\epsilon_1$ (top), $\epsilon_2$ (bottom). (left) Absolute token weight difference distribution between OTPO and DPO. (middle) The distribution normalization value $\tau$ in Eq.\ref{eq:rescale}. (right) Changes in reward margin. }
    \label{fig:hyperparameter_sensitivity}
\end{figure*}

We analyze the sensitivity of OTPO to hyperparameters $\epsilon_1$, $\epsilon_2$, focusing on their effects on absolute token weight difference, normalization value $\tau$, and reward margin. 
Llama-3.2-3B and the HelpSteer2 dataset are utilized for this study.
The results, presented in Fig.~\ref{fig:hyperparameter_sensitivity}, reveal several key insights.

\textbf{Impact of $\epsilon_1$ on token weight difference.}
Higher values of $\epsilon_1$ result in smaller weight differences, as the entropy regularization term imposes stronger penalization, leading to smoother weight distributions. On the other hand, $\epsilon_2$ shows almost no impact on token weight difference.

\textbf{Differing changes of $\epsilon_1, \epsilon_2$ on normalization value $\tau$.}
A larger $\epsilon_2$ decreases $\tau$ by imposing stronger penalties on marginal differences, whereas increasing $\epsilon_1$ implicitly downweights the impact of marginal differences overall, leading to a higher $\tau$. This interplay reflects the contrasting roles of these hyperparameters in shaping the weight normalization process, as well as the necessity of the normalization term $\tau$ in stabilizing training.

\textbf{Stability of reward margin}.
Despite substantial variations in absolute token weight difference and $\tau$,  reward margins exhibit remarkable stability, 
The reward margin remains stable at approximately 0.12 across different configurations, and is consistently higher than DPO significantly.
This stability can be attributed to the joint effects of Optimal Transport and weight sum normalization, which mitigate fluctuations in token weights while preserving the overall quality of reward estimation. Thus, OTPO ensures a stable reward estimation under varying hyperparameter configurations.

\section{Length Analysis}

In this section, we further analyze OTPO's effect by comparing the sensitivity of response log probability and reward difference to response length (difference).

\textbf{OTPO exhibits a relatively smaller positive relationship to length compared to DPO}, as shown in Fig.~\ref{fig:logp_length}. 
Specifically, the solid dots in the figure represent the actual test samples, while the dashed lines indicate linear fits based on these points. The gentler slope of OTPO relative to DPO suggests that OTPO provides more stable log probability estimates despite variations in response length. This aligns with the findings in Fig.~\ref{fig:teasor}, reinforcing OTPO’s ability to mitigate length bias.

\textbf{OTPO leads to more varied reward differences than DPO}, attributing to its contextual awareness, as shown in Fig.~\ref{fig:reward_length_diff}. Unlike DPO, where reward differences remain narrowly distributed around 0.1 regardless of length differences, OTPO exhibits a broader range of reward differences, from -0.4 to 1.0. Notably, responses with larger absolute reward differences are associated with minimal length differences, indicating that OTPO’s reward differences are driven primarily by contextual rather than length-related factors. This contextual sensitivity allows OTPO to dynamically optimize reward differences by focusing on meaningful content variations rather than superficial length differences. As a result, OTPO demonstrates superior adaptability and robustness in aligning reward optimization with instruction-following behavior.

\section{Case Study on the Transport Plan}
\label{sec:transport_plan}

We further analyze the transport plan $\Gamma$ visualized as the Sankey diagram in Fig.~\ref{fig:teasor}, providing insights into the intuition behind optimal transport. 
In this diagram, each token is represented by a node, with tokens from the chosen response on the left and those from the rejected response on the right. Each node is labeled with its position in the pairwise data, its token text, and its aggregate weight $\omega_*^i$. Token positions are indicated using response codes (C for chosen and R for rejected) and their indices, e.g., \textit{C4} represents the 4th token in the chosen response. The lines connecting the nodes in the middle illustrate the transport flow, $\Gamma_{i,j}$, between token pairs, where the thickness of the lines reflects the magnitude of the flow. The height of each node is proportional to its aggregate weight, which corresponds to the sum of its inflow or outflow. To highlight the distribution of weights, we divide tokens into terciles: tokens in the highest tercile are colored \textcolor{teal}{teal}, those in the middle tercile are \textcolor{light_blue}{light blue}, and those in the lowest tercile are \textcolor{orange}{orange}.

\textbf{Shared and semantically similar tokens are densely mapped together.} Tokens from the phrase ``The capital of France is Paris'' are assigned higher weights ($> 1.5$), as shown by the larger bars for these tokens and stronger connections between them. These parts directly address the question, making them critical for reward estimation. Conversely, less relevant tokens in the divergent parts, such as ``known for its art'' or ``the birthplace of Victor Hugo'' are assigned with lower weights ($< 0.7$), as they have thinner connections to other tokens and smaller bars.
These parts, while potentially informative, are less central to the core instruction and are consequently de-emphasized. 
This weighting strategy ensures that the reward difference focuses on the most meaningful content, enhancing the stability of reward estimation.

\textbf{A Smooth Transition of Weight}. 
The Sankey diagram also illustrates a smooth transition of token weights, as optimal transport gradually shifts the emphasis from shared, contextually important tokens to less relevant, unique tokens.
Despite the large weights ($> 1.5$) received by the upper shared part, other contextually important tokens receive moderately high weights ($0.8-0.9$), including the token ``city'' describing Paris, and the period marking the end of a sentence. This progression is evident in the intermediate-sized bars and connections associated with these tokens.
As the responses diverge semantically, the weights of unique tokens progressively diminish ($< 0.7$). Tokens like ``known for its art'' or ``the birthplace of Victor Hugo'' cannot be well-mapped to corresponding tokens in the other response and therefore spread out thinly across many connections, as shown by the sparse and diffused lines in the lower part of the figure. This smooth transition ensures a natural weighting scheme that reflects the semantic relevance of each token. 

\section{Connection to Existing Rewards}
\label{sec:reward_model}

We use leave-one-out to explain the importance of each token to the final reward signal produced by current reward models, and compare the generated explanation with OT weight.

\begin{table}[!t]
    \centering
    \begin{tabular}{cc}
    \toprule
    Method & Weight Visualization \\
    \midrule
      OT Weight   & \tabincell{l}{\sethlcolor{blue!52.33632}\hl{The} \sethlcolor{blue!57.78742}\hl{capital} \sethlcolor{blue!70.0}\hl{of} \sethlcolor{blue!36.56914}\hl{France} \sethlcolor{blue!34.247417}\hl{is} \sethlcolor{blue!56.080975}\hl{Paris} \sethlcolor{blue!38.193295}\hl{,} \sethlcolor{blue!36.52137}\hl{a} \\ \sethlcolor{blue!32.305813}\hl{major} \sethlcolor{blue!30.671368}\hl{European} \sethlcolor{blue!33.93367}\hl{city} \sethlcolor{blue!33.066948}\hl{known} \sethlcolor{blue!36.468407}\hl{for} \sethlcolor{blue!36.89949}\hl{its} \\ \sethlcolor{blue!31.755896}\hl{art} \sethlcolor{blue!34.81017}\hl{,} \sethlcolor{blue!35.913864}\hl{culture} \sethlcolor{blue!39.034687}\hl{,} \sethlcolor{blue!38.082794}\hl{and} \sethlcolor{blue!37.99243}\hl{history} \sethlcolor{blue!61.52185}\hl{.}} \\
      \hline 
      ArmoRM & \tabincell{l}{\sethlcolor{blue!33.409090909090914}\hl{The} \sethlcolor{blue!56.47727272727273}\hl{capital} \sethlcolor{blue!33.409090909090914}\hl{of} \sethlcolor{blue!53.29545454545455}\hl{France} \sethlcolor{blue!58.06818181818182}\hl{is} \sethlcolor{blue!70.0}\hl{Paris} \sethlcolor{blue!33.409090909090914}\hl{,} \sethlcolor{blue!17.5}\hl{a} \\ \sethlcolor{blue!10.340909090909092}\hl{major} \sethlcolor{blue!16.704545454545457}\hl{European} \sethlcolor{blue!27.045454545454543}\hl{city} \sethlcolor{blue!24.65909090909091}\hl{known} \sethlcolor{blue!23.863636363636363}\hl{for} \sethlcolor{blue!14.318181818181818}\hl{its}  \\ \sethlcolor{blue!11.136363636363637}\hl{art} \sethlcolor{blue!15.909090909090908}\hl{,} \sethlcolor{blue!15.909090909090908}\hl{culture} \sethlcolor{blue!15.909090909090908}\hl{,} \sethlcolor{blue!58.86363636363637}\hl{and} \sethlcolor{blue!33.409090909090914}\hl{history} \sethlcolor{blue!50.11363636363637}\hl{.}} \\
    \bottomrule
    \end{tabular}
    \caption{Token weight visualization, darker color denotes higher weight. 
    OT represents weights computed via Optimal Transport, while ArmoRM, DPO, and OTPO denotesthe naive leave-one-out explanations of reward prediction. 
    }
    \label{tab:rm_explain}
\end{table}


\textbf{Generating token-level explanations}. To compare the weighting scheme with existing rewards, we generate token-level explanations of explicit or implicit rewards. We applied a naive leave-one-out method, where each token was iteratively replaced with a padding token, and the resulting change in the reward score was measured. The proportional reduction in the reward score was treated as the token’s contribution. 

\textbf{Connection to explicit reward model}.
We observed a high correlation of 0.76 between token-level weights computed using Optimal Transport and the explanation of predictions by an external reward model, ArmoRM~\cite{wang2024interpretable}, as shown in Tab.~\ref{tab:rm_explain}. ArmoRM was selected for its compatibility with the tokenizer of the backbone model, Llama-3-8B, ensuring consistent tokenization. Both approaches focus strongly on the straightforward response to the prompt while downweighting more detailed explanations. Interestingly, since the rejected response contains a similar expression of the straightforward answer \textit{``The capital of France is Paris''}, the OT weighting scheme also emphasizes this part. This aligns with the intuition behind OTPO, which prioritizes shared components that are more likely to be relevant to the prompt.

\begin{figure}
    \centering
    \includegraphics[width=0.95\linewidth]{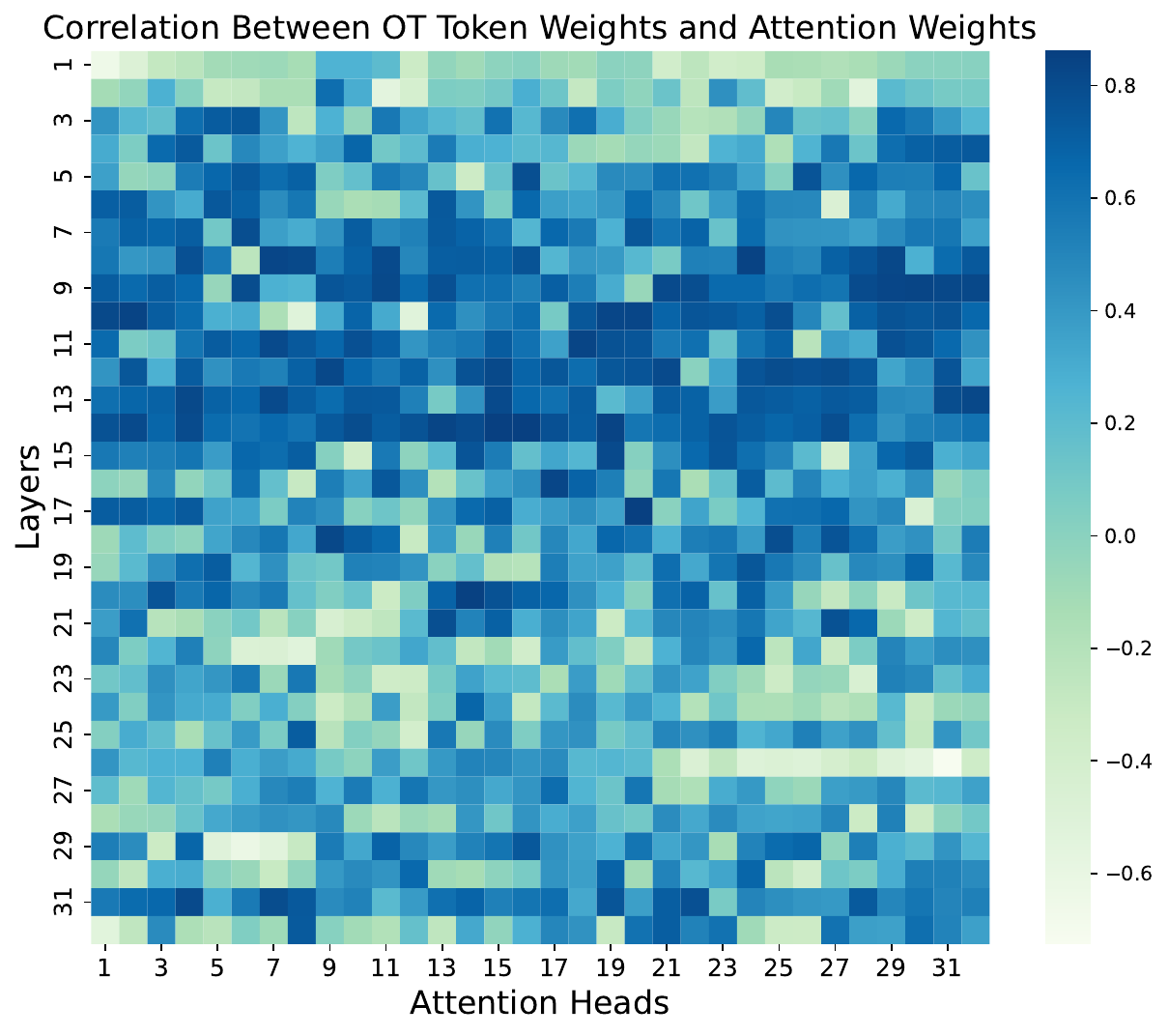}
    \caption{Spearman correlation between token weighting scheme by Optimal Transport and the last token attention weights of ArmoRM at each layer and attention head.}
    \label{fig:ot_attn_corr}
\end{figure}

\textbf{Relationship between OT weights and attention weights}. We further examine the correlation between token-level weights derived from Optimal Transport and the attention weights of ArmoRM, as illustrated in Fig.~\ref{fig:ot_attn_corr}. The highest correlations are predominantly found in the middle layers, where the model aggregates sequential context and refines its understanding of the input. This observation aligns with prior findings that model explanations often rely heavily on the middle layers~\cite{kim2018interpretability}, where meaningful internal representations emerge. Interestingly, despite lower correlations between OT weights and attention weights in the upper and lower layers, the leave-one-out explanations in Tab.~\ref{tab:rm_explain} still exhibit strong overall agreement with OT. This implies that the middle layers are critical in bridging the gap between token-level contributions and global model explanations.

\section{Evaluation Configuration Details}

\subsection{Summarization Win Rate Calculation}

In this section, we include the details for GPT-4o to generate win rates for summarization. The order of summaries is randomly chosen for every evaluation. Considering the special characteristic of \textbf{short} for summaries, we prompt the model to generate less than 48 words, and treat the summaries with more than 48 words as ``Lose''. Below is the prompt for GPT-4o.

\begin{verbatim}
Which of the following summaries does a 
better job of summarizing the most im-
-portant points in the given forum post, 
without including unimportant or ir-
-relevant details? A good summary is 
both precise and concise.

Post:
{post}

Summary A:
{summary_A}

Summary B:
{summary_B}

FIRST provide a one-sentence comparison 
of the two summaries, explaining which 
you prefer and why. SECOND, on a new 
line, state only "A" or "B" to indi-
-cate your choice.

Your response should use the format:
Comparison: <one-sentence comparison 
and explanation>
Preferred: <"A" or "B">
\end{verbatim}

\subsection{Human Evaluation}
\label{sec:human_evaluation_detail}

In our human evaluation, we recruited human experts to choose the better response among two responses to the same question according to certain requirements. The human experts possess a high school level of English proficiency, allowing for easy comprehension of the responses. These experts were selected from within our academic institution to ensure a consistent educational background. To maintain the quality of annotation, we implemented a compensation structure that rewards the experts based on the number of pairwise responses they annotate. This approach was designed to incentivize thorough and careful consideration of each response pair.

During the evaluation, the experts were required to complete their assessments within a two-minute window for each response. This time constraint was established to simulate a realistic scenario in which users need to make quick judgments about the preference of responses. Both of the experts were presented with the same set of 50 pairwise responses for each method, totaling 250 pairwise responses, to ensure consistency in the evaluation process. 

Given the two responses to the same question, the evaluator needs to compare them from five perspectives, including relevance, coherence, completeness, conciseness, and instruction following. Then, the evaluator should make a judgment on which is the better response (winning response), or if the two responses are equally good/bad. The winning response will receive a score of 1, while the losing response will receive a score of 0. When the two responses are considered equally good/bad, then they both receive a score of 0.5.

\begin{figure*}
    \centering
    \includegraphics[width=0.99\linewidth]{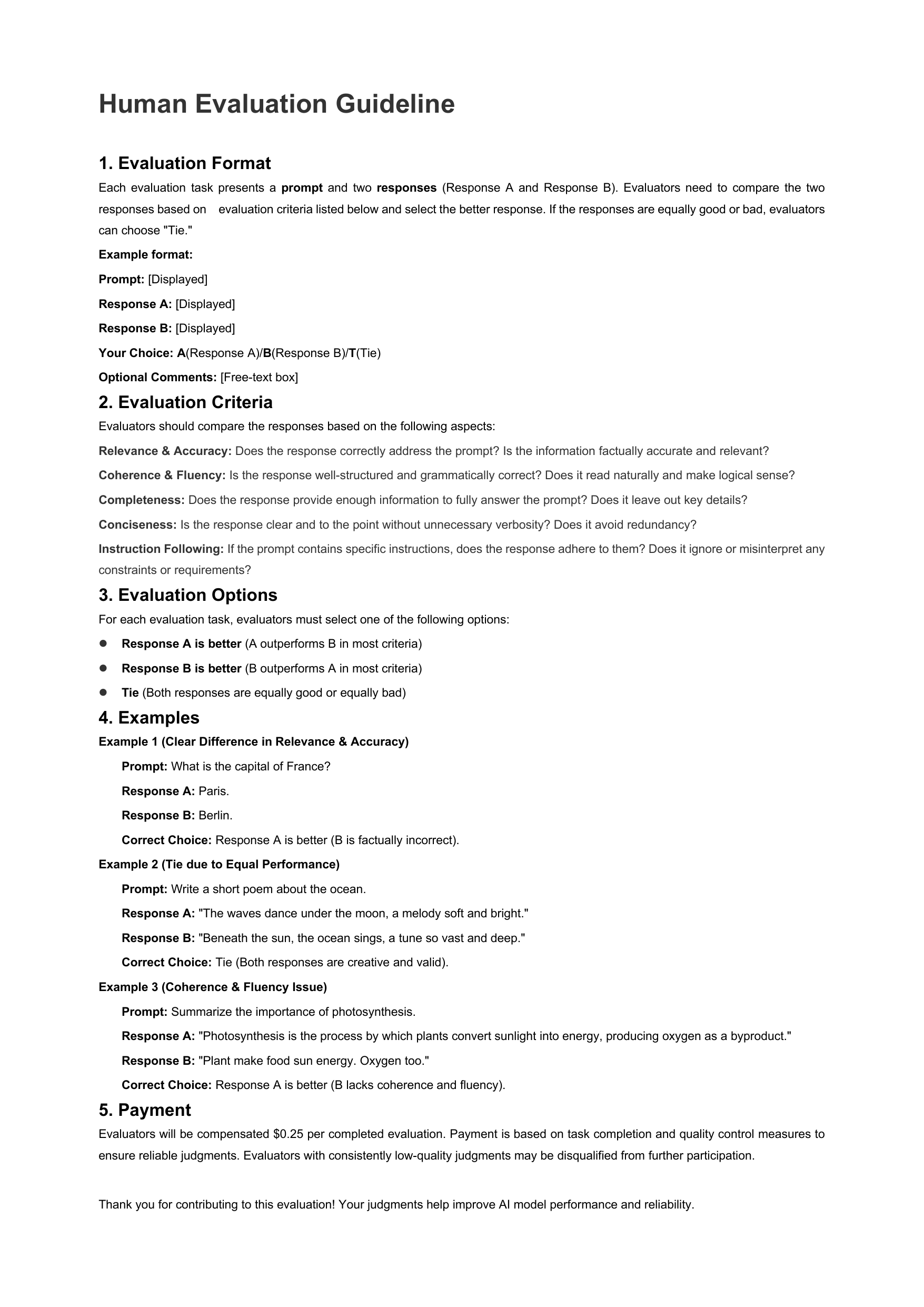}
    \caption{Human Evaluation Guideline.}
    \label{fig:human_evaluation_guideline}
\end{figure*}

We show guidelines that were provided to the evaluators in Fig.~\ref{fig:human_evaluation_guideline}. These guidelines were crafted to assist the evaluators in their task and to standardize the evaluation criteria.

\onecolumn

\section{Mathematical Derivations}

\subsection{Token-level DPO loss}
\label{app:token_dpo_loss}

We can transform a response $y$'s probability $\pi(y|x)$ given x as follows:
\begin{align}
    \pi(y|x) & = \prod_{i=1}^{|y|} \pi(y^i | y^{< i}, x) 
         = \exp(\log\prod_{i=1}^{|y|} \pi(y^i | y^{< i}, x)) 
         = \exp(\sum_{i=1}^{|y|} \log\pi(y^i | y^{< i}, x))
        \label{eq:probability_token}
\end{align}
For the reward difference term in Eq.~\ref{eq:dpo_reward}:
\begin{align}
    \Delta_r & = \log\frac{\pi_{\theta}(y_c|x)}{\pi_{\text{ref}}(y_c|x)} 
    -  \log\frac{\pi_{\theta}(y_r|x)}{\pi_{\text{ref}}(y_r|x)} 
\end{align}
Since the likelihood of a response is modeled as the multiplicative probability of each token:
\begin{equation}
    \pi_{*}(y|x) = \prod_{i=1}^{|y|} \pi_*(y^i | y^{< i}, x)
\end{equation}
We can express $\Delta_r$ as:
\begin{align}
        \Delta_r & = \log\frac{\prod_{i=1}^{|y_c|} \pi_{\theta}(y_c^i | y_c^{< i}, x)}{\prod_{i=1}^{|y_c|} \pi_{\text{ref}}(y_c^i | y_c^{< i}, x)} 
 \quad -  \log\frac{\prod_{j=1}^{|y_r|} \pi_{\theta}(y_r^j | y_r^{< j}, x)}{\prod_{j=1}^{|y_r|} \pi_{\text{ref}}(y_r^j | y_r^{< j}, x)}
     \\
        & = \sum_{i=1}^{|y_c|} \log\frac{ \pi_{\theta}(y_c^i | y_c^{< i}, x)}{\pi_{\text{ref}}(y_c^i | y_c^{< i}, x)}   \quad -  \sum_{j=1}^{|y_r|}\log\frac{ \pi_{\theta}(y_r^j | y_r^{< j}, x)}{ \pi_{\text{ref}}(y_r^j | y_r^{< j}, x)} 
\end{align}

\subsection{Token Weight Derivation}

We provide the derivations of the results in Tab.~\ref{tab:weight_baselines}.

\textbf{SimPO}. The loss in \cite{meng2024simpo} can be transformed as:
\begin{align}
    \mathcal{L}_{simpo} & = -\log \sigma  \left( \frac{\beta}{|y_c|} \log \pi_\theta(y_c|x) - \frac{\beta}{|y_r|} \log \pi_\theta(y_r|x) - \gamma \right) \\
    & = -\log \sigma  \left( \frac{\beta}{|y_c|} \sum_{i=1}^{|y_c|} \log\pi(y_c^i | y_c^{< i}, x)) - \frac{\beta}{|y_r|} \sum_{j=1}^{|y_r|} \log\pi(y_r^j | y_r^{< j}, x) - \gamma \right) \\
    & = -\log \sigma  \left( \beta \sum_{i=1}^{|y_c|} \frac{1}{|y_c|}\log\pi(y_c^i | y_c^{< i}, x)) - \beta \sum_{j=1}^{|y_r|}\frac{1}{|y_r|} \log\pi(y_r^j | y_r^{< j}, x) - \gamma \right) 
\end{align}
Here, $\gamma$ controls the overall margin to be optimized.

\textbf{SamPO}. Assume the shorter response's length is $m = min(|y_c|, |y_r|)$, the loss in \cite{lu2024eliminating} can be transformed as:
\begin{align}
    \mathcal{L}_{SamPO} = -\log \sigma \left( \beta \sum_{t=1}^{m}\log \frac{\pi_\theta(y_c^t|x)}{\pi_{\text{ref}}(y_c^t|x)} - \beta \sum_{t=1}^{m} \log \frac{\pi_\theta(y_r^m|x)}{\pi_{\text{ref}}(y_r^m|x)}\right), \;
     \text{where} \; y^t \sim \text{Uniform}(m, {y}^T)
\end{align}
Here $\text{Uniform}(m, {y}^T)$ denotes uniformly sample $m$ tokens from all tokens in response $y$. Moving the sampling operation to the token index, we have:
\begin{align}
    \mathcal{L}_{SamPO} = -\log \sigma \left( \beta \sum_{t\in S_c}\log \frac{\pi_\theta(y_c^t|x, y_c^{<t})}{\pi_{\text{ref}}(y_c^t|x, y_c^{<t})} - \beta \sum_{t\in S_r} \log \frac{\pi_\theta(y_r^t|x, y_r^{<t})}{\pi_{\text{ref}}(y_r^t|x, y_r^{<t})}\right), \; \\
 \text{where} \; S_* \sim \text{Uniform}(m, [1, |y_*|])
\end{align}
Here $\text{Uniform}(m, [1, |y_*|])$ denotes uniformly sample $m$ numbers from all integers from 1 to $|y_*|$.

\textbf{LDDPO}. \cite{liu2024length} transforms the probability of response to:
\begin{align}
    \hat{\pi}(y|x) & = \prod_{i=1}^m \pi(y^i|x, y^{<i}) \prod_{i=m+1}^{|y|} \pi^{\alpha}(y^i|x, y^{<i}) \\
    & = \exp(\log\prod_{i=1}^m \pi(y^i|x, y^{<i}) \prod_{i=m+1}^{|y|} \pi^{\alpha}(y^i|x, y^{<i})) \\
    & = \exp(\sum_{i=1}^{m} \log\pi(y^i | y^{< i}, x) + \sum_{i=m+1}^{|y|} \alpha \log\pi(y^i | y^{< i}, x))
\end{align}
Thus the loss can be derived as:
\begin{align}
    \mathcal{L}_{LDDPO} = & -\log \sigma \left( \beta \log \frac{\hat{\pi}_\theta(y_c|x)}{\hat{\pi}_{\text{ref}}(y_c|x)} - \beta \ \log \frac{\hat{\pi}_\theta(y_r|x)}{\hat{\pi}_{\text{ref}}(y_r|x)}\right) \\
    \begin{split}
        = & -\log \sigma ( \beta (\sum_{t=1}^{m}\log \frac{\pi_\theta(y_c^t|x)}{\pi_{\text{ref}}(y_c^t|x)} + \alpha \sum_{t=m+1}^{|y_c|}\log \frac{\pi_\theta(y_c^t|x)}{\pi_{\text{ref}}(y_c^t|x)})
        \\
        & - \beta (\sum_{t=1}^{m} \log \frac{\pi_\theta(y_r^t|x)}{\pi_{\text{ref}}(y_r^t|x)} + \alpha \sum_{t=m+1}^{|y_r|} \log \frac{\pi_\theta(y_r^t|x)}{\pi_{\text{ref}}(y_r^t|x)} ) ), \; m = min(|y_c|, |y_r|) 
    \end{split}
\end{align}

\subsection{Gradient Analysis}

The derivative for the log-sigmoid function is:
\begin{align}
    \frac{\delta \log \sigma (u)}{\delta u} & = \frac{1}{\sigma(u)}\frac{\delta \sigma(u)}{\delta u} 
    = \frac{1}{\sigma(u)} (\sigma(u) (1- \sigma(u))) 
    = (1- \sigma(u)) 
    = \sigma(-u)
    \label{eq:log_sigma_derivative}
\end{align}
We can derive the gradient of OTPO as:
\begin{align}
    \nabla \mathcal{L}_{\text{OTPO}}(\pi_{\theta}) = -\beta\mathbb{E}_{ D}\sigma(-\beta \Delta_{\hat{r}}) \nabla (\Delta_{\hat{r}}) 
    \label{eq:otwpo_overall}
\end{align}
\begin{equation}
        \Delta_{\hat{r}} = \sum_{i=1}^{|y_c|} \omega_c^{*i}\log \frac{ \pi_{\theta}(y_c^i|x, y_c^{< i})}{\pi_{\text{ref}}(y_c^i|x, y_c^{< i})}
        - \sum_{i=1}^{|y_r|} \omega_r^{*i}\log \frac{ \pi_{\theta}(y_r^i|x, y_r^{< i})}{\pi_{\text{ref}}(y_r^i|x, y_r^{< i})}
    \label{eq:otwpo_delta_r}
\end{equation}
\begin{equation}
        \nabla(\Delta_{\hat{r}}) = \sum_{i=1}^{|y_c|} \omega_c^{*i}\nabla_{\theta}\log \pi_{\theta}(y_c^i|x, y_c^{< i})
        - \sum_{i=1}^{|y_r|} \omega_r^{*i} \nabla_{\theta} \log \pi_{\theta}(y_r^i|x, y_r^{< i})
    \label{eq:otwpo_delta_gradient}
\end{equation}
Similarly, the gradient of DPO is:
\begin{align}
    \nabla \mathcal{L}_{\text{DPO}}(\pi_{\theta}) = -\beta\mathbb{E}_{ D}\sigma(-\beta \Delta_{r}) \nabla (\Delta_{r}) 
    \label{eq:dpo_overall}
\end{align}
\begin{equation}
        \Delta_{r} = \sum_{i=1}^{|y_c|} \log \frac{ \pi_{\theta}(y_c^i|x, y_c^{< i})}{\pi_{\text{ref}}(y_c^i|x, y_c^{< i})}
        - \sum_{i=1}^{|y_r|} \log \frac{ \pi_{\theta}(y_r^i|x, y_r^{< i})}{\pi_{\text{ref}}(y_r^i|x, y_r^{< i})}
    \label{eq:dpo_delta_r}
\end{equation}
\begin{equation}
        \nabla(\Delta_{r}) = \sum_{i=1}^{|y_c|} \nabla_{\theta}\log \pi_{\theta}(y_c^i|x, y_c^{< i})
        - \sum_{i=1}^{|y_r|} \nabla_{\theta} \log \pi_{\theta}(y_r^i|x, y_r^{< i})
    \label{eq:dpo_delta_gradient}
\end{equation}

We compare the gradients of OTPO and DPO to understand the impact on training and alignment results. For the reward difference term in Eq.~\ref{eq:otwpo_delta_r}, \ref{eq:dpo_delta_r}, OTPO is relatively smaller and more stable, as those semantically dissimilar tokens are down-weighted. While for gradient updates in Eq.~\ref{eq:otwpo_delta_gradient}, \ref{eq:dpo_delta_gradient}, the gradient scale of OTPO on each token is additionally controlled by the OT weighting scheme instead of uniform update, performing larger updates on the more important tokens related to prompt and the other response, while downweighting the updates in the less relevant tokens. This ensures a more meaningful and concentrated gradient update compared to DPO. Overall, the OT weighting scheme ensures more stable reward difference term and dynamic gradient updates given the context and pairwise data information.

\end{document}